\documentclass{article}
\usepackage[final, nonatbib]{neurips_2021}
\usepackage[numbers]{natbib}
\usepackage{fancyhdr}
\usepackage{color}
\usepackage[ruled]{algorithm2e}

\usepackage[utf8]{inputenc} 
\usepackage[T1]{fontenc}    
\usepackage[colorlinks=true, citecolor=blue, linkcolor=black]{hyperref}     
\usepackage{url}            
\usepackage{booktabs}       
\usepackage{amsfonts}       
\usepackage{nicefrac}       
\usepackage{microtype}      
\usepackage{xcolor}         

\usepackage{microtype}
\usepackage{graphicx}
\usepackage{wrapfig}
\usepackage{subcaption}
\usepackage{booktabs} 
\usepackage{amsthm}
\usepackage{amsmath}
\usepackage{mathtools}
\mathtoolsset{showonlyrefs=true}
\usepackage{bbold}
\usepackage{hyperref}

\newtheorem{definition}{Definition}
\newtheorem{remark}{Remark}
\newtheorem{lemma}{Lemma}

\newtheorem{proposition}{Proposition}
\newtheorem{theorem}{Theorem}

\usepackage{nth}
\newcommand{\bracks}[2]{\underbracket{#1 \dots #1}_{#2}}

\title{There Is No Turning Back:\\ A Self-Supervised Approach for\\Reversibility-Aware Reinforcement Learning}
\author{
Nathan Grinsztajn\thanks{Equal contribution.}\\
Inria, Scool Team\\
CRIStAL, CNRS, Université de Lille\\
\texttt{nathan.grinsztajn@inria.fr}\\
\And
Johan Ferret\footnotemark[1]\\
Google Research, Brain Team\\
Inria, Scool Team\\
CRIStAL, CNRS, Université de Lille\\
\And
Olivier Pietquin\\
Google Research, Brain Team\\
\And
Philippe Preux\\
Inria, Scool Team\\
CRIStAL, CNRS, Université de Lille\\
\And
Matthieu Geist\\
Google Research, Brain Team\\
}
\date{}

\begin{document}

\maketitle

\begin{abstract}
We propose to learn to distinguish reversible from irreversible actions for better informed decision-making in Reinforcement Learning (RL). From theoretical considerations, we show that approximate reversibility can be learned through a simple surrogate task: ranking randomly sampled trajectory events in chronological order. Intuitively, pairs of events that are always observed in the same order are likely to be separated by an irreversible sequence of actions. 
Conveniently, learning the temporal order of events can be done in a fully self-supervised way, which we use to estimate the reversibility of actions from experience, without any priors.
We propose two different strategies that incorporate reversibility in RL agents, one strategy for exploration (RAE) and one strategy for control (RAC). We demonstrate the potential of reversibility-aware agents in several environments, including the challenging Sokoban game. In synthetic tasks, we show that we can learn control policies that never fail and reduce to zero the side-effects of interactions, even without access to the reward function. 
\end{abstract}



\section{Introduction}

We address the problem of estimating if and how easily actions can be reversed in the Reinforcement Learning (RL) context.
Irreversible outcomes are often not to be taken lightly when making decisions. As humans, we spend more time evaluating the outcomes of our actions when we know they are irreversible~\citep{mcallister1979}. As such, irreversibility can be positive (\textit{i.e.} takes risk away for good) or negative (\textit{i.e.} leads to later regret). Also, decision-makers are more likely to anticipate regret for hard-to-reverse decisions~\citep{zeelenberg1999}. All in all, irreversibility seems to be a good prior to exploit for more principled decision-making. In this work, we explore the option of using irreversibility to guide decision-making and confirm the following assertion: by estimating and factoring reversibility in the action selection process, safer behaviors emerge in environments with intrinsic risk factors. In addition to this, we show that exploiting reversibility leads to more efficient exploration in environments with undesirable irreversible behaviors, including the famously difficult Sokoban puzzle game.

\begin{wrapfigure}{r}{0.5\textwidth}
  \vspace{-20pt}
  \includegraphics[width=\linewidth]{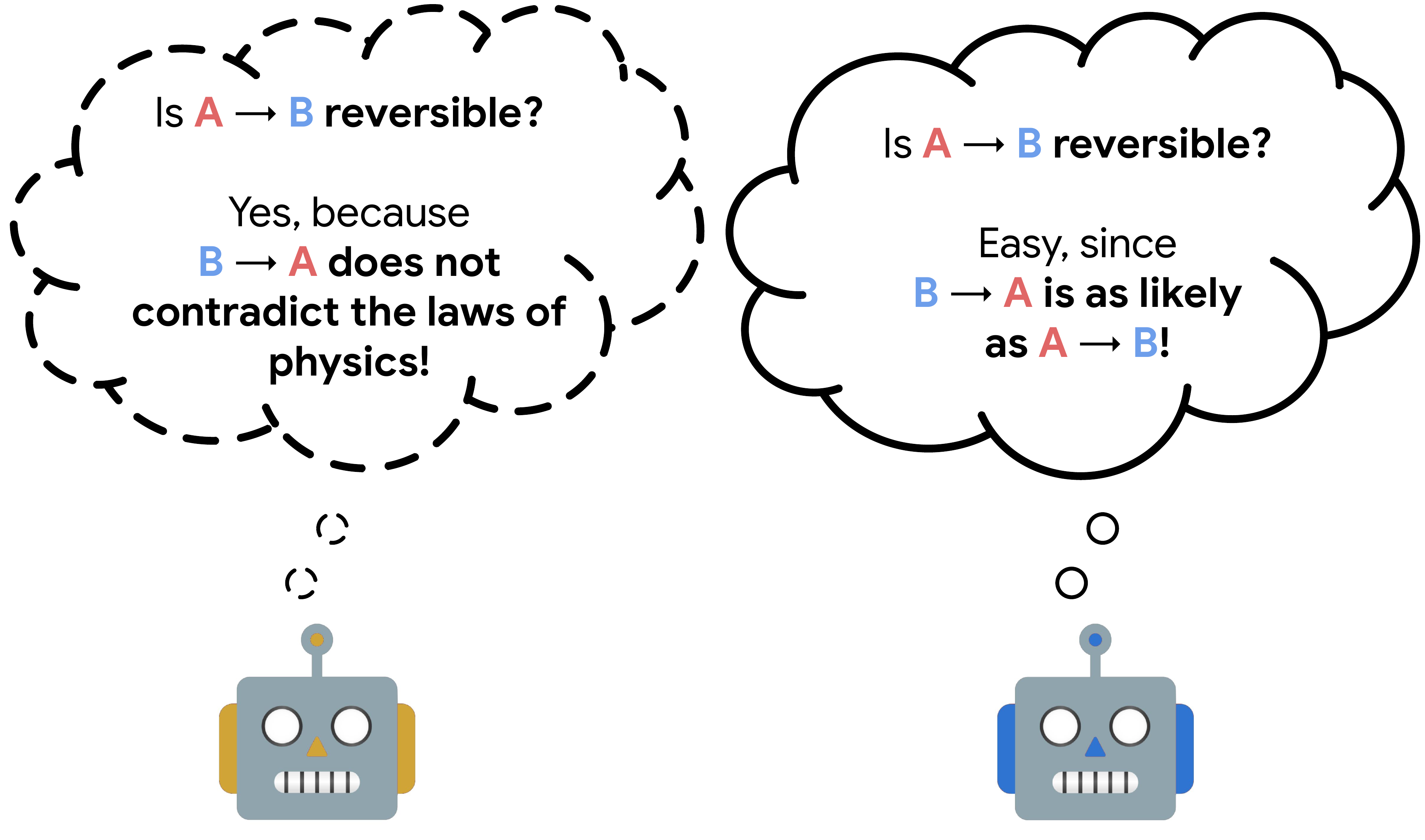}
  \caption{High-level illustration of how reversibility can be estimated. \textbf{Left:} from an understanding of physics. \textbf{Right:} ours, from experience.}
  \label{fig:intuition}
  \vspace{-15pt}
\end{wrapfigure}

However, estimating the reversibility of actions is no easy feat. It seemingly requires a combination of planning and causal reasoning in large dimensional spaces. We instead opt for another, simpler approach (see Fig.~\ref{fig:intuition}): we propose to learn in which direction time flows between two observations, directly from the agents’ experience, and then consider \textit{irreversible} the transitions that are assigned a temporal direction with high confidence. \textit{In fine}, we reduce reversibility to a simple classification task that consists in predicting the temporal order of events.

\looseness=-1
Our contributions are the following: 1) we formalize the link between reversibility and precedence estimation, and show that reversibility can be approximated via temporal order, 2) we propose a practical algorithm to learn temporal order in a self-supervised way, through simple binary classification using sampled pairs of observations from trajectories, 3) we propose two novel exploration and control strategies that incorporate reversibility, and study their practical use for directed exploration and safe RL, illustrating their relative merits in synthetic as well as more involved tasks such as Sokoban puzzles.



\section{Related Work}

\looseness=-1
To the best of our knowledge, this work is the first to explicitly model the reversibility of transitions and actions in the context of RL, using temporal ordering to learn from trajectories in a self-supervised way, in order to guide exploration and control. Yet, several aspects of the problem we tackle were studied in different contexts, with other motivations; we review these here. 

\textbf{Leveraging reversibility in RL.}
\citet{kruusmaa2007} estimate the reversibility of state-action couples so that robots avoid performing irreversible actions, since they are more likely to damage the robot itself or its environment. A shortcoming of their approach is that they need to collect explicit state-action pairs and their reversal actions, which makes it hard to scale to large environments. 
Several works~\citep{savinov2019, badia2020never, badia2020agent57} use reachability as a curiosity bonus for exploration: if the current state has a large estimated distance to previous states, it means that it is novel and the agent should be rewarded. Reachability and reversibility are related, in the sense that irreversible actions lead to states from which previous states are unreachable. Nevertheless, their motivations and ours diverge, and we learn reversibility through a less involved task than that of learning reachability. 
\citet{nair2020} learn to reverse trajectories that start from a goal state so as to generate realistic trajectories that reach similar goals. In contrast, we use reversibility to direct exploration and/or control, not for generating learning data.
Closest to our work,~\citet{rahaman2020} propose to learn a potential function of the states that increases with time, which can detect irreversibility to some extent. A drawback of the approach is that the potential function is learned using trajectories sampled from a random policy, which is a problem for many tasks where a random agent might fail to cover interesting parts of the state space. In comparison, our method does not use a potential function and learns jointly with the RL agent, which makes it a viable candidate for more complex tasks.

\paragraph{Safe exploration.}
Safe exploration aims at making sure that the actions of RL agents do not lead to negative or unrecoverable effects that would outweigh the long-term value of exploration~\citep{Amodei2016}.
Notably, previous works developed distinct approaches to avoid irreversible behavior: by incremental updates to safe policies~\citep{Hans2008, garcia2012}, which requires knowing such a policy in advance; by restricting policy search to ergodic policies~\citep{Moldovan2012} (\textit{i.e.} that can always come back to any state visited), which is costly; by active exploration~\citep{maillard2019}, where the learner can ask for rollouts instead of exploring potentially unsafe areas of the state space itself; and by computing regions of attraction~\citep{Berkenkamp2016} (the part of the state space where a controller can bring the system back to an equilibrium point), which requires prior knowledge of the environment dynamics.

\paragraph{Self-supervision from the arrow of time.}
Self-supervision has become a central component of modern machine learning algorithms, be it for computer vision, natural language or signal processing. In particular, using temporal consistency as a source of self-supervision is now ubiquitous, be it to learn representations for downstream tasks~\citep{goroshin2015, ramanathan2015, dadashi2020}, or to learn to detect temporal inconsistencies~\citep{wei2018}.
The closest analogies to our work are methods that specifically estimate some aspects of the arrow of time as self-supervision. Most are to be found in the video processing literature, and self-supervised tasks include predicting which way the time flows~\citep{pickup2014, wei2018}, verifying the temporal order of a subset of frames~\citep{misra2016}, predicting which video clip has the wrong temporal order among a subset~\citep{fernando2017} as well as reordering shuffled frames or clips from the video~\citep{fernando2015, el2019, xu2019}. \citet{bai2020} notably propose to combine several of these pretext tasks along with data augmentation for video classification. Using time as a means of supervision was also explored for image sequencing~\citep{basha2012}, audio~\citep{carr2021} or EEG processing~\citep{saeed2020}.
In RL, self-supervision also gained momentum in recent years~\citep{guo2020, srinivas2020, yarats2021}, with temporal information being featured~\citep{amiranashvili2018}. 
Notably, several works~\citep{aytar2018, dwibedi2018, guo2018, sermanet2018} leverage temporal consistency to learn useful representations, effectively learning to discriminate between observations that are temporally close and observations that are temporally distant.  
In comparison to all these works, we estimate the arrow of time through temporal order prediction with the explicit goal of finding irreversible transitions or actions.

\section{Reversibility} 


\paragraph{Degree of Reversibility.} 
\label{section:definition_reversibility}
We start by introducing formally the notion of reversibility. Intuitively, an action is reversible if it can be undone, meaning that there is a sequence of actions that can bring us back to the original state. 



\begin{definition}
\label{definition:degree_of_rev}
Given a state $s$, we call \emph{degree of reversibility within $K$ steps} of an action $a$
\begin{equation}
\phi_K(s, a) \coloneqq  \sup_{\pi} p_\pi(s \in \tau_{t+1:t+K+1}\mid s_t = s, a_t=a),
\end{equation}
and the \emph{degree of reversibility} of an action is defined as
\begin{equation}
\phi(s, a) 
\coloneqq \sup_{\pi} p_\pi(s \in \tau_{t+1:\infty}\mid s_t = s, a_t=a),
\end{equation}
\looseness=-1
with $\tau = \{ s_i \}_{i=1 \, \ldots \, T} \sim \pi$ corresponding to a trajectory, and $\tau_{t:t'}$ the subset of the trajectory between the timesteps $t$ and $t'$ (excluded). We omit their dependency on $\pi$ for the sake of conciseness.
%
Given $s \in S$, the action $a$ is \emph{reversible} if and only if $\phi(s, a) = 1$, and said \emph{irreversible} if and only if $\phi(s, a) = 0$.
\end{definition}

In deterministic environments, an action is either reversible or irreversible: given a state-action couple $(s, a)$ and the unique resulting state $s'$, $\phi_K(s, a)$ is equal to 1 if there is a sequence of less than $K$ actions which brings the agent from $s'$ to $s$, and is otherwise equal to zero. In stochastic environments, a given sequence of actions can only reverse a transition up to some probability, hence the need for the notion of degree of reversibility.


\paragraph{Policy-Dependent Reversibility.} 
In practice, it is useful to quantify the degree of reversibility of an action as the agent acts according to a fixed policy $\pi$, for which we extend the notions introduced above. We simply write :
\begin{equation}
\phi_{\pi, K}(s, a)  \coloneqq p_\pi(s \in \tau_{t+1:t+K+1}\mid s_t = s, a_t=a) \text{ and }
\phi_\pi(s, a)  \coloneqq p_\pi(s \in \tau_{t+1:\infty}\mid s_t = s, a_t=a).
\end{equation}

It immediately follows that $\phi_K(s, a) = \sup_{\pi} \phi_{\pi, K}(s, a)$ and $\phi(s, a) = \sup_{\pi} \phi_\pi(s, a)$.

\section{Reversibility Estimation via Classification}
\label{sec:ReversibilityEstimation_via_Classification}

Quantifying the exact degree of reversibility of actions is generally hard.
In this section, we show that reversibility can be approximated efficiently using simple binary classification.

\subsection{Precedence Estimation}
Supposing that a trajectory contains the states $s$ and $s'$, we want to be able to establish \textit{precedence}, that is predicting whether $s$ or $s'$ comes first \textit{on average}. It is a binary classification problem, which consists in estimating the quantity $\mathbb{E}_{s_t=s, s_{t'} = s'} \big[ \mathbb{1}_{t'>t} \big] $.
%
%
Accordingly, we introduce the precedence estimator which, using a set of trajectories, learns to predict which state of an arbitrary pair is most likely to come first.

\begin{definition}
Given a fixed policy $\pi$, we define the \emph{finite-horizon precedence estimator} between two states as follows:
\begin{equation}
    {\psi}_{\pi, T}(s, s') = \mathbb{E}_{\tau \sim \pi} \, \mathbb{E}_{\substack{s_{t}=s, s_{t'}=s' \\ t, t' < T}} \big[ \mathbb{1}_{t'>t} \big].
\end{equation}
\end{definition}

Conceptually, given two states $s$ and $s'$, the precedence estimator gives an approximate probability of $s'$ being visited after $s$, given that both $s$ and $s'$ are observed in a trajectory. The indices are sampled uniformly within the specified horizon $T \in \mathbb{N}$, so that this quantity is well-defined even for infinite trajectories. Additional properties of $\psi$, regarding transitivity for instance, can be found in Appx.~\ref{appendix:extra_properties}.
\begin{remark}
The quantity $\psi_{\pi, T}(s, s')$ is only defined for pairs of states which can be found in the same trajectory, and is otherwise irrelevant. In what follows, we implicitly impose this condition when considering state pairs.
\end{remark}

\begin{theorem}
\label{theorem:convergence}
For every policy $\pi$ and $s, s' \in S$, $\psi_{\pi, T}(s, s')$ converges when $T$ goes to infinity. We refer to the limit as the \emph{precedence estimator}, written $\psi_\pi(s, s')$.
\end{theorem}

The proof of this theorem is developed in Appendix~\ref{appendix:proofs_both_theorem}. This result is key to ground theoretically the notion of empirical reversibility $\Bar{\phi}$, which we introduce in the next definition. It simply consists in extending the notion of precedence to a state-action pair.

\begin{definition}
We finally define the \emph{empirical reversibility} using the precedence estimator:
\begin{align}
{\Bar{\phi}}_\pi(s, a) &=\mathbb{E}_{s' \sim P(s, a)}\big[{\psi}_\pi(s', s)\big]. 
\end{align}
\end{definition}
In a nutshell, given that we start in $s$ and take the action $a$, the empirical reversibility $\Bar{\phi}_\pi(s, a)$ measures the probability that we go back to $s$, starting from a state $s'$ that follows $(s, a)$. 
We now show that our empirical reversibility is linked with the notion of reversibility defined in the previous section, and can behave as a useful proxy.

\subsection{Estimating Reversibility from Precedence}

We present here our main theoretical result which relates reversibility and empirical reversibility:
\begin{theorem}
\label{theorem:inequality_over_2}
Given a policy $\pi$, a state $s$ and an action $a$, we have: $\Bar{\phi}_\pi(s, a) \geq \frac{\phi_\pi(s, a)}{2} $.
\end{theorem}
The full proof of the theorem is given in Appendix~\ref{appendix:proofs_both_theorem}. 

This result theoretically justifies the name of empirical reversibility. From a practical perspective, it provides a way of using $\Bar{\phi}$ to detect actions which are irreversible or hardly reversible: $\Bar{\phi}_\pi(s, a) \ll 1$ implies $\phi_\pi(s, a)\ll 1$ and thus provides a sufficient condition to detect actions with low degrees of reversibility.
This result gives a way to detect actions that are irreversible given a specific policy followed by the agent. Nevertheless, we are generally interested in knowing if these actions are irreversible for any policy, meaning $\phi(s,a) \ll 1$ with the definition of Section~\ref{section:definition_reversibility}. The next proposition makes an explicit connection between $\Bar{\phi}_\pi$ and $\phi$, under the assumption that the policy $\pi$ is stochastic.


\begin{proposition}
We suppose that we are given a state $s$, an action $a$ such that $a$ is reversible in $K$ steps, and a policy $\pi$. Under the assumption that $\pi$ is stochastic enough, meaning that there exists $\rho > 0$ such that for every state and action $s, a$, $\pi(a \mid s) > \rho$, we have:   
$\Bar{\phi}_\pi(s, a) \geq \frac{\rho^K}{2} $. Moreover, we have for all $K \in \mathbb{N}$:
$\Bar{\phi}_\pi(s, a) \geq \frac{\rho^K}{2} \phi_K(s, a)$.
\end{proposition}

The proof is given in Appendix~\ref{appendix:prop1}. As before, this proposition gives a practical way of detecting irreversible moves. If for example $\Bar{\phi}_\pi(s, a) < {\rho^k}/{2}$ for some $k \in \mathbb{N}$, we can be sure that action $a$ is not reversible in $k$ steps. The quantity $\rho$ can be understood as a minimal probability of taking any action in any state. This condition is not very restrictive:  $\epsilon$-greedy strategies for example satisfy this hypothesis with $\rho = \frac{\epsilon}{\lvert A \rvert}$.

In practice, it can also be useful to limit the maximum number of time steps between two sampled states. That is why we also define the windowed precedence estimator as follows: 
\begin{definition}
Given a fixed policy $\pi$, we define the \emph{windowed precedence estimator} between two states as follows:
%
\begin{equation}
    {\psi}_{\pi, T, w}(s, s') = \mathbb{E}_{\tau \sim \pi} \mathbb{E}_{\substack{s_{t}=s, s_{t'}=s' \\ t, t' <T \\ \vert t-t' \vert \leq w}} \big[ \mathbb{1}_{t'>t} \big].
\end{equation}
\end{definition}

Intuitively, compared to previous precedence estimators, ${\psi}_{\pi, T, w}$ is restricted to short-term dynamics, which is a desirable property in tasks where distinguishing the far future from the present is either trivial or impossible.

\section{Reversibility-Aware Reinforcement Learning}

\begin{figure}
  \centering
  \includegraphics[width=\textwidth]{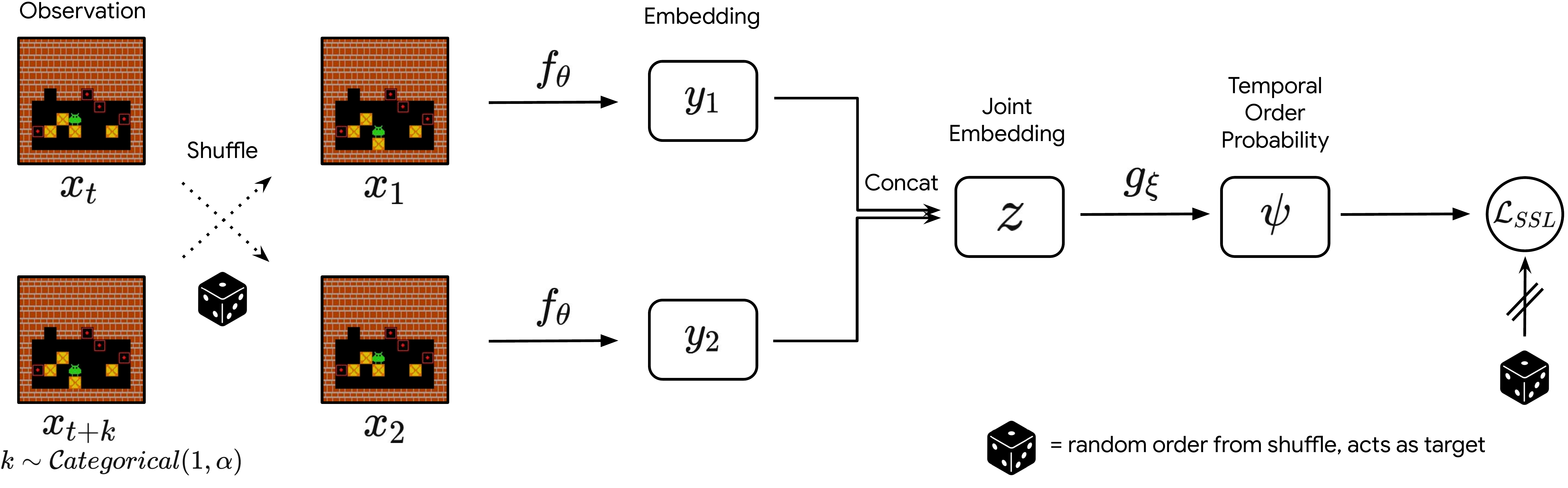}
  \caption{The proposed self-supervised procedure for precedence estimation.}
  \label{fig:precedence-estimation}
\end{figure}

Leveraging the theoretically-grounded bridge between precedence and reversibility established in the previous section, we now explain how reversibility can be learned from the agent’s experience and used in a practical setting. 

\paragraph{Learning to rank events chronologically.}
Learning which observation comes first in a trajectory is achieved by binary supervised classification, from pairs of observations sampled uniformly in a sliding window on observed trajectories. This can be done fully offline, \textit{i.e.} using a previously collected dataset of trajectories for instance, or fully online, \textit{i.e.} jointly with the learning of the RL agent; but also anywhere on the spectrum by leveraging variable amounts of offline and online data.

\label{par:limitations}
This procedure is not without caveats. In particular, we want to avoid overfitting to the particularities of the behavior of the agent, so that we can learn meaningful, generalizable statistics about the order of events in the task at hand. Indeed, if an agent always visits the state $s_a$ before $s_b$, the classifier will probably assign a close-to-one probability that $s_a$ precedes $s_b$. This might not be accurate with other agents equipped with different policies, unless transitioning from $s_b$ to $s_a$ is hard due to the dynamics of the environment, which is in fact exactly the cases we want to uncover. We make several assumptions about the agents we apply our method to: 1) agents are learning and thus, have a policy that changes through interactions in the environment, 2) agents have an incentive not to be too deterministic. For this second assumption, we typically use an entropic regularization in the chosen RL loss, which is a common design choice in modern RL methods. These assumptions, when put together, alleviate the risk of overfitting to the idiosyncrasies of a single, non-representative policy. 

We illustrate the precedence classification procedure in Fig.~\ref{fig:precedence-estimation}. 
A temporally-ordered pair of observations, distant of no more than $w$ timesteps, is sampled from a trajectory and uniformly shuffled. The result of the shuffling operation is memorized and used as a target for the binary classification task. A Siamese network creates separate embeddings for the pair of observations, which are concatenated and fed to a separate feed-forward network, whose output is passed through a sigmoid to obtain a probability of precedence. This probability is updated via negative log-likelihood against the result of the shuffle, so that it matches the actual temporal order.

Then, a transition (and its implicit sequence of actions) represented by a starting observation $x$ and a resulting observation $x'$ is deemed irreversible if the estimated precedence probability $\psi(x, x')$ is superior to a chosen threshold $\beta$.
Note that we do not have to take into account the temporal proximity of these two observations here, which is a by-product of sampling observations uniformly in a window in trajectories.
Also, depending on the threshold $\beta$, we cover a wide range of scenarios, from pure irreversibility ($\beta$ close to $1$) to soft irreversibility ($\beta > 0.5$, the bigger $\beta$, the harder the transition is to reverse).
This is useful because different tasks call for different levels of tolerance for irreversible behavior: while a robot getting stuck and leading to an early experiment failure is to be avoided when possible, tasks involving human safety might call for absolute zero tolerance for irreversible decision-making. We elaborate on these aspects in Sec.~\ref{section:experiments}.

\begin{figure}
  \centering
  \includegraphics[width=\textwidth]{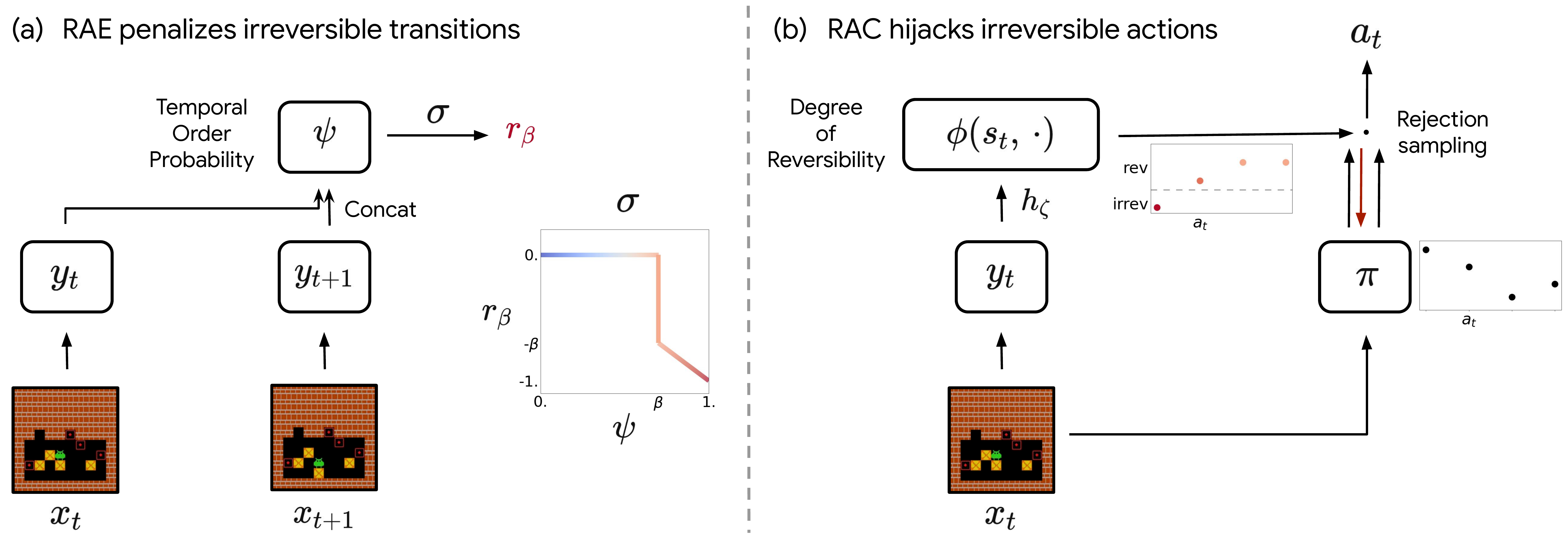}
  \caption{Our proposed methods for reversibility-aware RL. \textbf{(a):} RAE encourages reversible behavior via auxiliary rewards. \textbf{(b):} RAC avoids irreversible behavior by rejecting actions whose estimated reversibility is inferior to a threshold.}
  \label{fig:rae-and-rac}
\end{figure}

\paragraph{Reversibility-Aware Exploration and Control.}
We propose two different algorithms based on reversibility estimation: Reversibility-Aware Exploration (RAE) and Reversibility-Aware Control (RAC). We give a high-level representation of how the two methods operate in Fig.~\ref{fig:rae-and-rac}. 

In a nutshell, RAE consists in using the estimated reversibility of a pair of consecutive observations to create an auxiliary reward function. In our experiments, the reward function is a piecewise linear function of the estimated reversibility and a fixed threshold, as in Fig.~\ref{fig:rae-and-rac}: it grants the agent a negative reward if the transition is deemed too hard to reverse. The agent optimizes the sum of the extrinsic and auxiliary rewards. Note that the specific function we use penalizes irreversible transitions but could encourage such transitions instead, if the task calls for it.

RAC can be seen as the action-conditioned counterpart of RAE. From a single observation, RAC estimates the degree of reversibility of all available actions, and “takes control” if the action sampled from the policy is not reversible enough (\textit{i.e.} has a reversibility inferior to a threshold $\beta$). “Taking control” can have many forms. In practice, we opt for rejection sampling: we sample from the policy until an action that is reversible enough is sampled. This strategy has the advantage of avoiding irreversible actions entirely, while trading-off pure reversibility for performance when possible. RAC is more involved than RAE, since the action-conditioned reversibility is learned from the supervision of a standard, also learned precedence estimator. Nevertheless, our experiments show that it is possible to learn both estimators jointly, at the cost of little overhead.

We now discuss the relative merits of the two methods.
In terms of applications, we argue that RAE is more suitable for directed exploration, as it only encourages reversible behavior. As a result, irreversible behavior is permitted if the benefits (\textit{i.e.} rewards) outweigh the costs (\textit{i.e.} irreversibility penalties). In contrast, RAC shines in safety-first, real-world scenarios, where irreversible behavior is to be banned entirely. With an optimal precedence estimator and task-dependent threshold, RAC will indeed hijack all irreversible sampled actions. RAC can be especially effective when pre-trained on offline trajectories: it is then possible to generate fully-reversible, safe behavior from the very first online interaction in the environment. We explore these possibilities experimentally in Sec.~\ref{subsection:reversible-policies}.    
   
Both algorithms can be used online or offline with small modifications to their overall logic. The pseudo-code for the online version of RAE and RAC can be found in Appendix~\ref{appendix:pseudo-code}.

The self-supervised precedence classification task could have applications beyond estimating the reversibility of actions: it could be used as a means of getting additional learning signal or representational priors for the RL algorithm. Nevertheless, we opt for a clear separation between the reversibility and the RL components so that we can precisely attribute improvements to the former, and leave aforementioned studies for future work. 



\section{Experiments}
\label{section:experiments}

The following experiments aim at demonstrating that the estimated precedence $\psi$ is a good proxy for reversibility, and at illustrating how beneficial reversibility can be in various practical cases. We benchmark RAE and RAC on a diverse set of environments, with various types of observations (tabular, pixel-based), using neural networks for function approximation. See Appendix~\ref{appendix:experiment_details} for details.

\subsection{Reward-Free Reinforcement Learning}
\label{subsection:reward_free}

We illustrate the ability of RAE to learn sensible policies without access to rewards.
We use the classic pole balancing task Cartpole \citep{cartpole}, using the OpenAI Gym~\citep{gym_openai} implementation.
In the usual setting, the agent gets a reward of 1 at every time step, such that the total undiscounted episode reward is equal to the episode length, and incentivizes the agent to learn a policy that stabilizes the pole.
Here, instead, we remove this reward signal and give a PPO agent~\citep{schulman2017} an intrinsic reward based on the estimated reversibility, which is learned online from agent trajectories. The reward function penalizes irreversibility, as shown in Fig.~\ref{fig:rae-and-rac}. Note that creating insightful rewards is quite difficult: too frequent negative rewards could lead the agent to try and terminate the episode as soon as possible. 

\begin{figure}
\centering
\begin{subfigure}{.33\textwidth}
  \centering
  \includegraphics[width=1\linewidth]{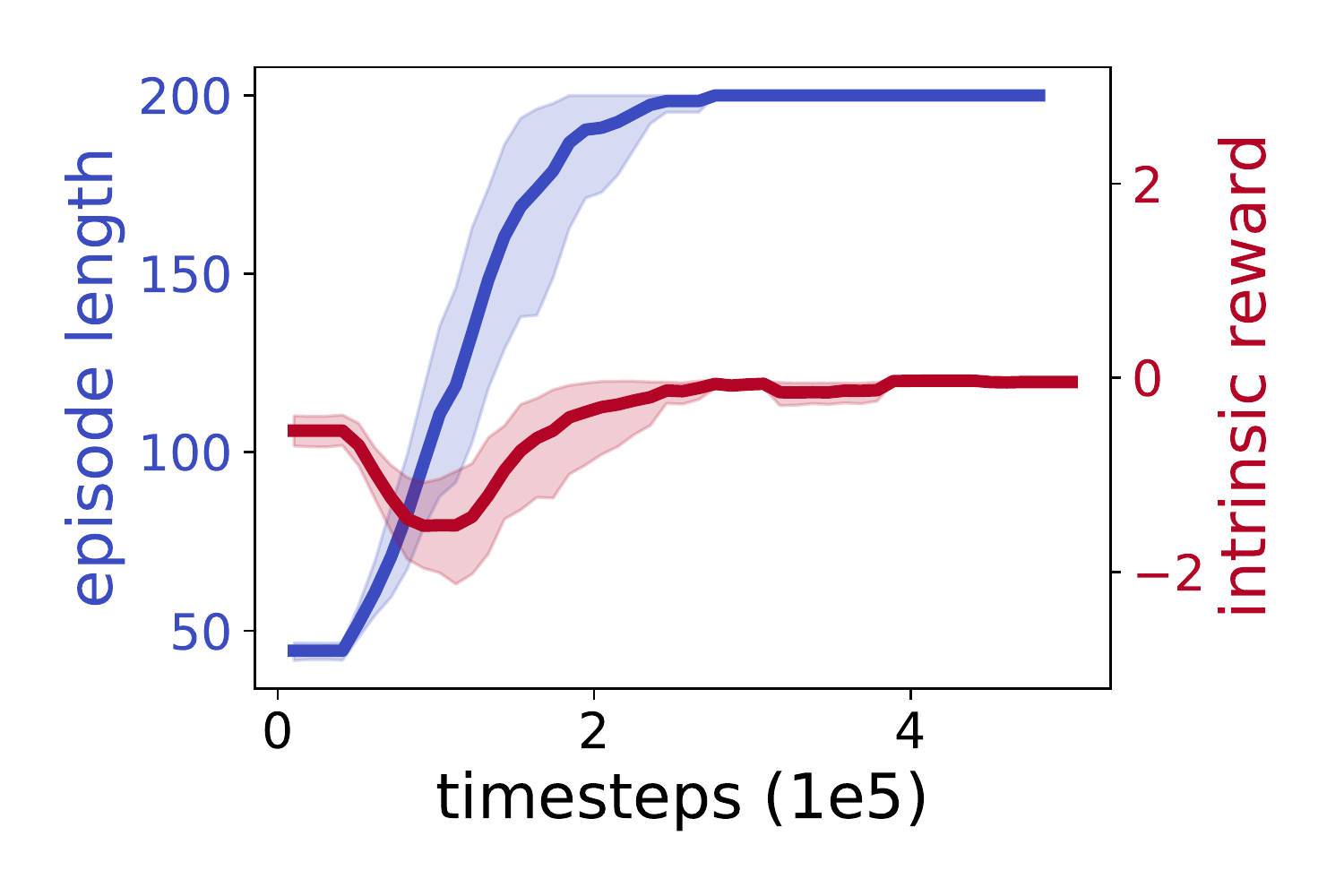}
  \caption{Training curves}
  \label{fig:reward_free_training_curves}
\end{subfigure}%
\begin{subfigure}{.33\textwidth}
  \centering
  \includegraphics[width=1\linewidth]{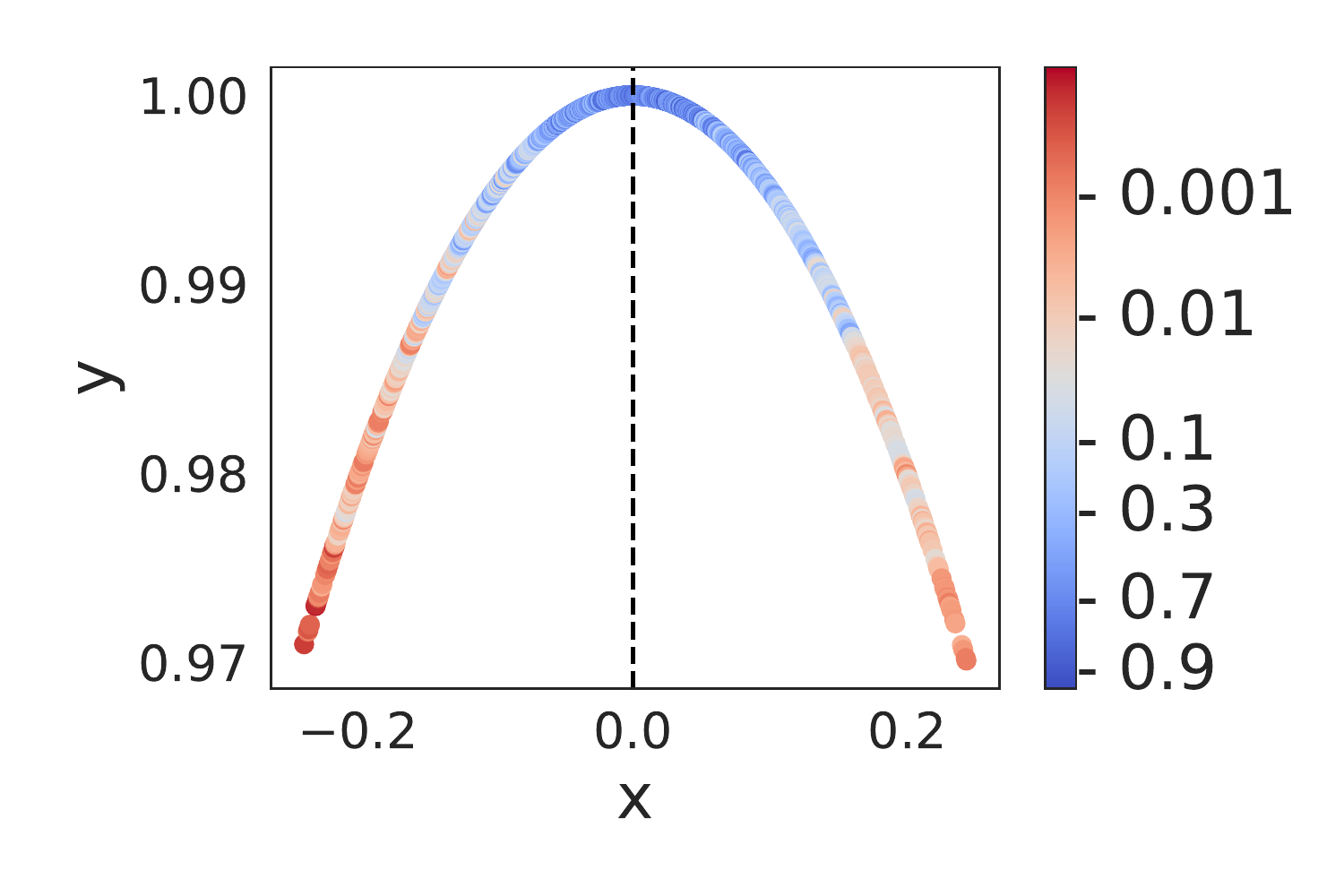}
  \caption{Relative pole coordinates}
  \label{fig:reward_free_x_y}
\end{subfigure}%
\begin{subfigure}{.33\textwidth}
  \centering
  \includegraphics[width=1\linewidth]{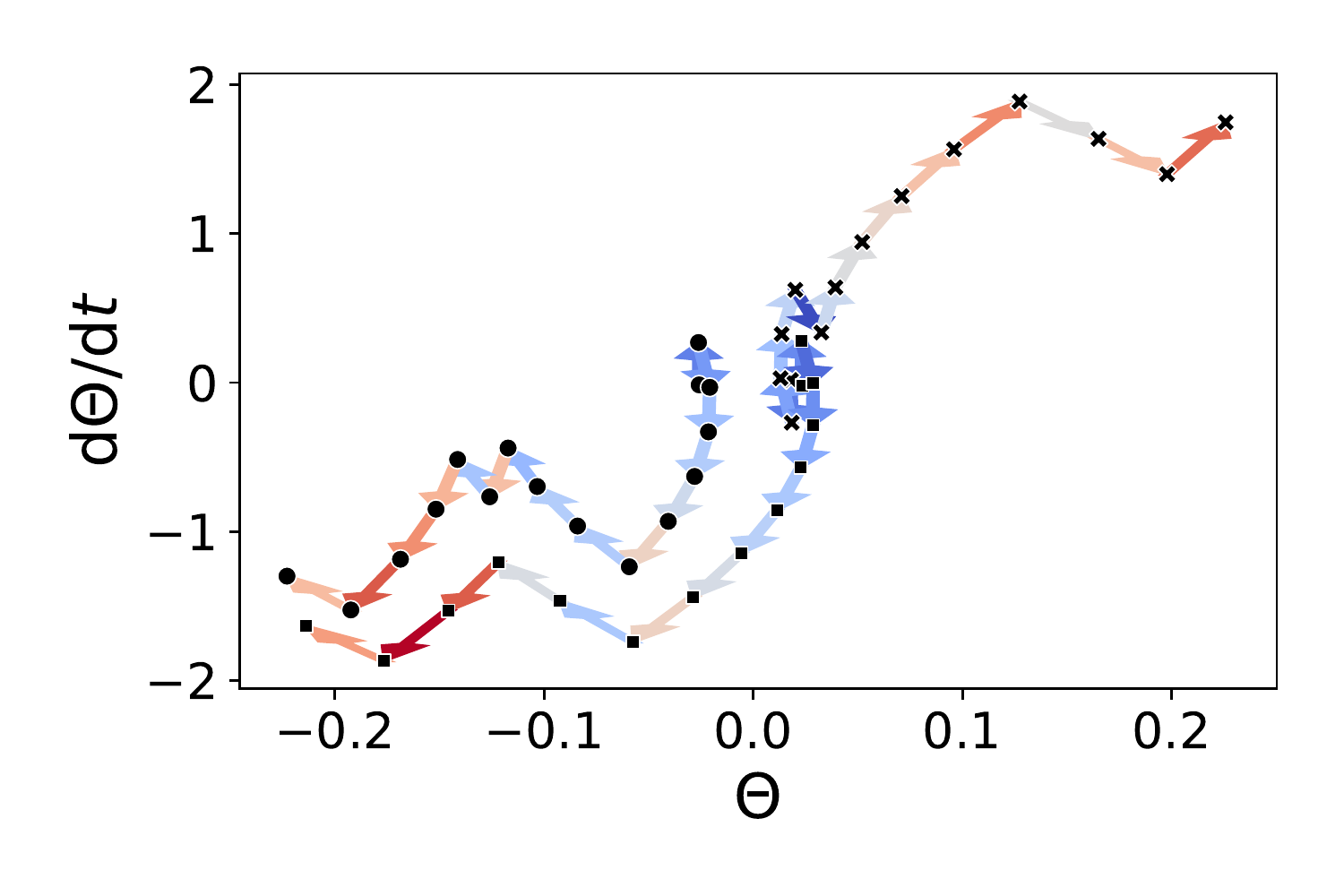}
  \caption{Random trajectories}
  \label{fig:reward_free_trajs}
\end{subfigure}
\caption{\textbf{(a):} Training curves of a PPO+RAE agent in reward-free Cartpole. Blue: episode length. Red: intrinsic reward. A 95\% confidence interval over 10 random seeds is shown. \textbf{(b):} The $x$ and $y$ axes are the coordinates of the end of the pole relatively to the cart position. The color denotes the online reversibility estimation between two consecutive states (logit scale). \textbf{(c):} The representation of three random trajectories according to $\theta$ (angle of the pole) and $\frac{d \theta}{d t}$. Arrows are colored according to the learned reversibility of the transitions they correspond to.}
\label{fig:cartpole_reward_free}
\end{figure}

We display our results in Fig.~\ref{fig:cartpole_reward_free}. Fig.~\ref{fig:reward_free_training_curves} confirms the claim that RAE can be used to learn meaningful rewards. Looking at the intrinsic reward, we discern three phases. Initially, both the policy and the reversibility classifier are untrained (and intrinsic rewards are 0). In the second phase, the classifier is fully trained but the agent still explores randomly (intrinsic rewards become negative). Finally, the agent adapts its behavior to avoid penalties (intrinsic rewards go to 0, and the length of trajectories increases). Our reward-free agent reaches the score of 200, which is the highest possible score.

To further assess the quality of the learned reversibility, we freeze the classifier after 300k timesteps and display its predicted probabilities according to the relative coordinates of the end of the pole (Fig.~\ref{fig:reward_free_x_y}) and the dynamics of the angle of the pole $\theta$ (Fig.~\ref{fig:reward_free_trajs}). In both cases, the empirical reversibility matches our intuition: the reversibility should decrease as the angle or angular momentum increase, since these coincide with an increasing difficulty to go back to the equilibrium.

\subsection{Learning Reversible Policies}
\label{subsection:reversible-policies}


In this section, we investigate how RAE can be used to learn reversible policies. When we train an agent to achieve a goal, we usually want it to achieve that goal following implicit safety constraints. Handcrafting such safety constraints would be time-consuming, difficult to scale for complex problems, and might lead to reward hacking; so a reasonable proxy consists in limiting irreversible side-effects in the environment \citep{Leike2017}.

\begin{figure}
\centering
\begin{subfigure}{.2\textwidth}
  \centering
  \includegraphics[width=1\linewidth]{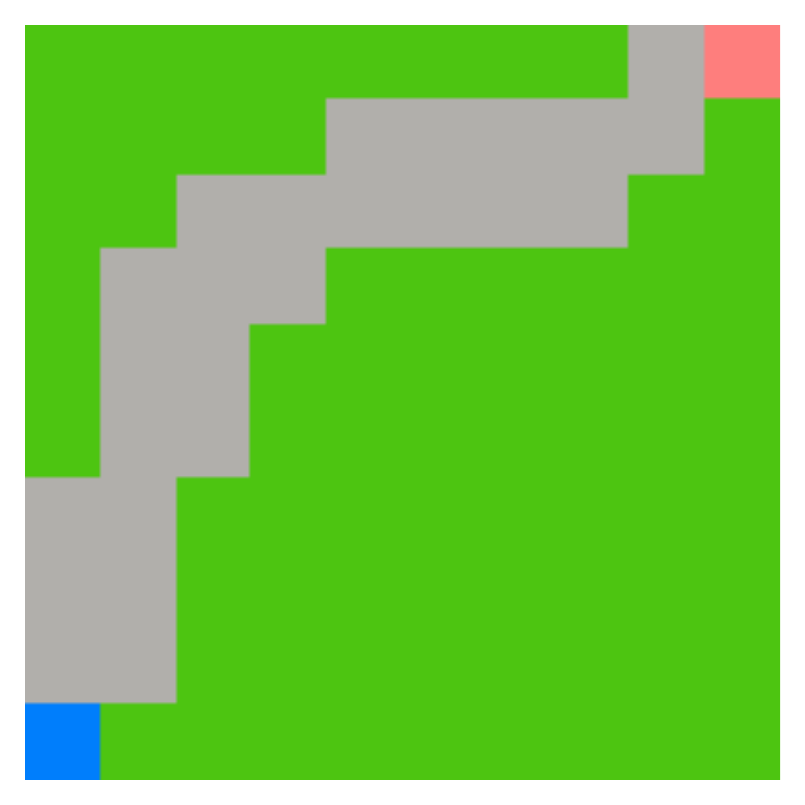}
  \caption{Initial state}
  \label{fig:grassland_init}
\end{subfigure}%
\begin{subfigure}{.2\textwidth}
  \centering
  \includegraphics[width=1\linewidth]{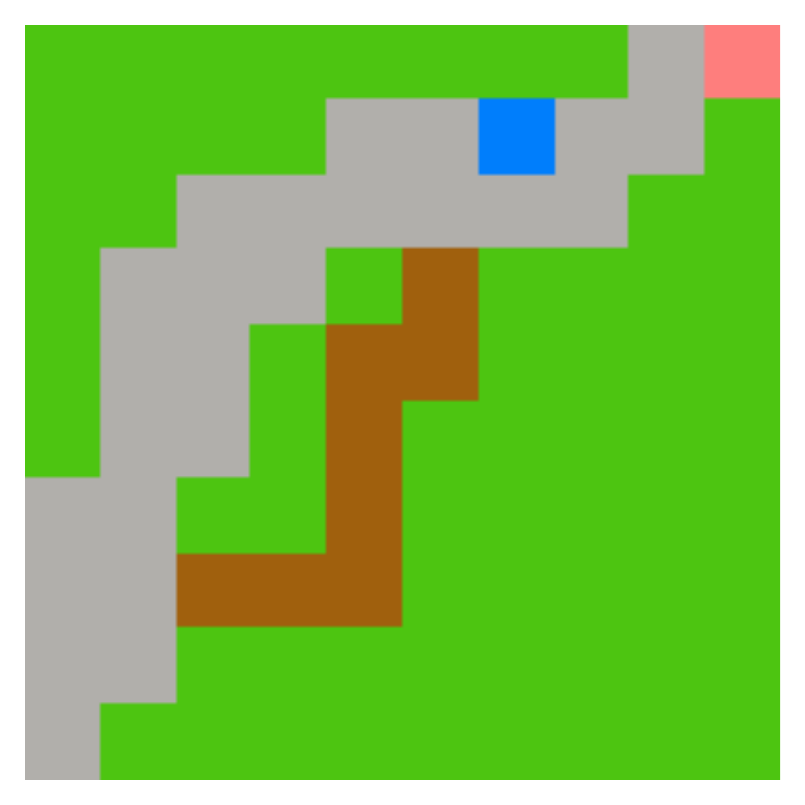}
  \caption{A trajectory}
  \label{fig:grassland_traj}
\end{subfigure}%
\begin{subfigure}{.2\textwidth}
  \centering
  \includegraphics[width=1\linewidth]{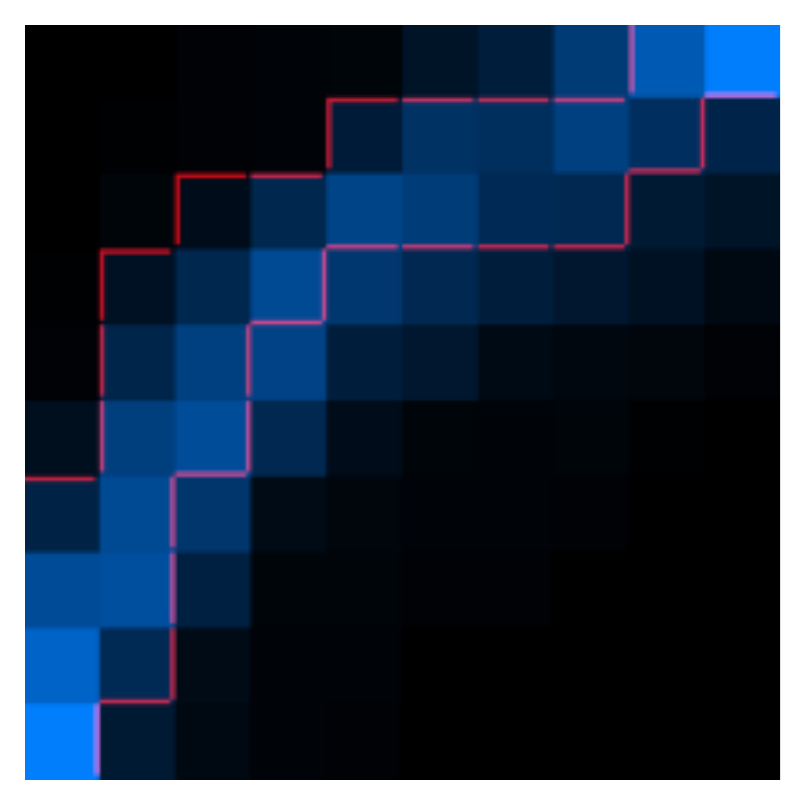}
  \caption{PPO (500k)}
  \label{fig:grassland_baseline}
\end{subfigure}%
\begin{subfigure}{.2\textwidth}
  \centering
  \includegraphics[width=1\linewidth]{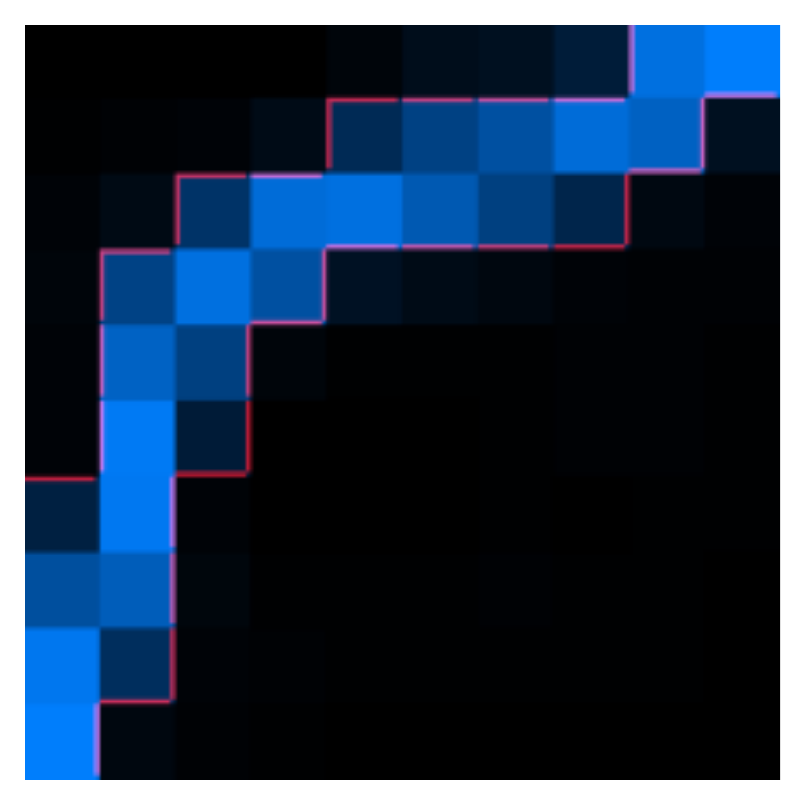}
  \caption{PPO+RAE (500k)}
  \label{fig:grassland_agent}
\end{subfigure}%
\caption{\textbf{(a):} The Turf environment. The agent can walk on grass, but the grass then turns brown. \textbf{(b):} An illustrative trajectory where the agent stepped on grass pixels. \textbf{(c):} State visitation heatmap for PPO. \textbf{(d):} State visitation heatmap for PPO+RAE. It coincides with the stone path (red).}
\label{fig:grassland}
\end{figure}

To quantify side-effects, we propose Turf, a new synthetic environment. As depicted in Fig.~\ref{fig:grassland_init},\ref{fig:grassland_traj}, the agent (blue) is rewarded when reaching the goal (pink). Stepping on grass (green) will spoil it, causing it to turn brown. Stepping on the stone path (grey) does not induce any side-effect.

In Fig.~\ref{fig:grassland_baseline},\ref{fig:grassland_agent}, we compare the behaviors of a trained PPO agent with and without RAE. The baseline agent is indifferent to the path to the goal, while the agent benefitting from RAE learns to follow the road, avoiding irreversible consequences.

\subsection{Sokoban}
\label{subsection:sokoban}

Sokoban is a popular puzzle game where a warehouse keeper (controlled by the player) must move boxes around and place them in dedicated places. Each level is unique and involves planning, since there are many ways to get stuck. For instance, pushing a box against a wall is often un-undoable, and prevents the completion of the level unless actually required to place the box on a specific target. 
Sokoban is a challenge to current model-free RL algorithms, as advanced agents require millions of interactions to reliably solve a fraction of levels~\citep{weber2017, guez2018}. One of the reasons for this is tied to exploration: since agents learn from scratch, there is a long preliminary phase where they act randomly in order to explore the different levels. During this phase, the agent will lock itself in unrecoverable states many times, and further exploration is wasted. It is worth recalling that contrary to human players, the agent does not have the option to reset the game when stuck. 
In these regards, Sokoban is a great testbed for reversibility-aware approaches, as we expect them to make the exploration phase more efficient, by incorporating the prior that irreversible transitions are to be avoided if possible, and by providing tools to identify such transitions.

\begin{figure}
\centering
\begin{subfigure}{.26\textwidth}
  \centering
  \includegraphics[width=1\linewidth]{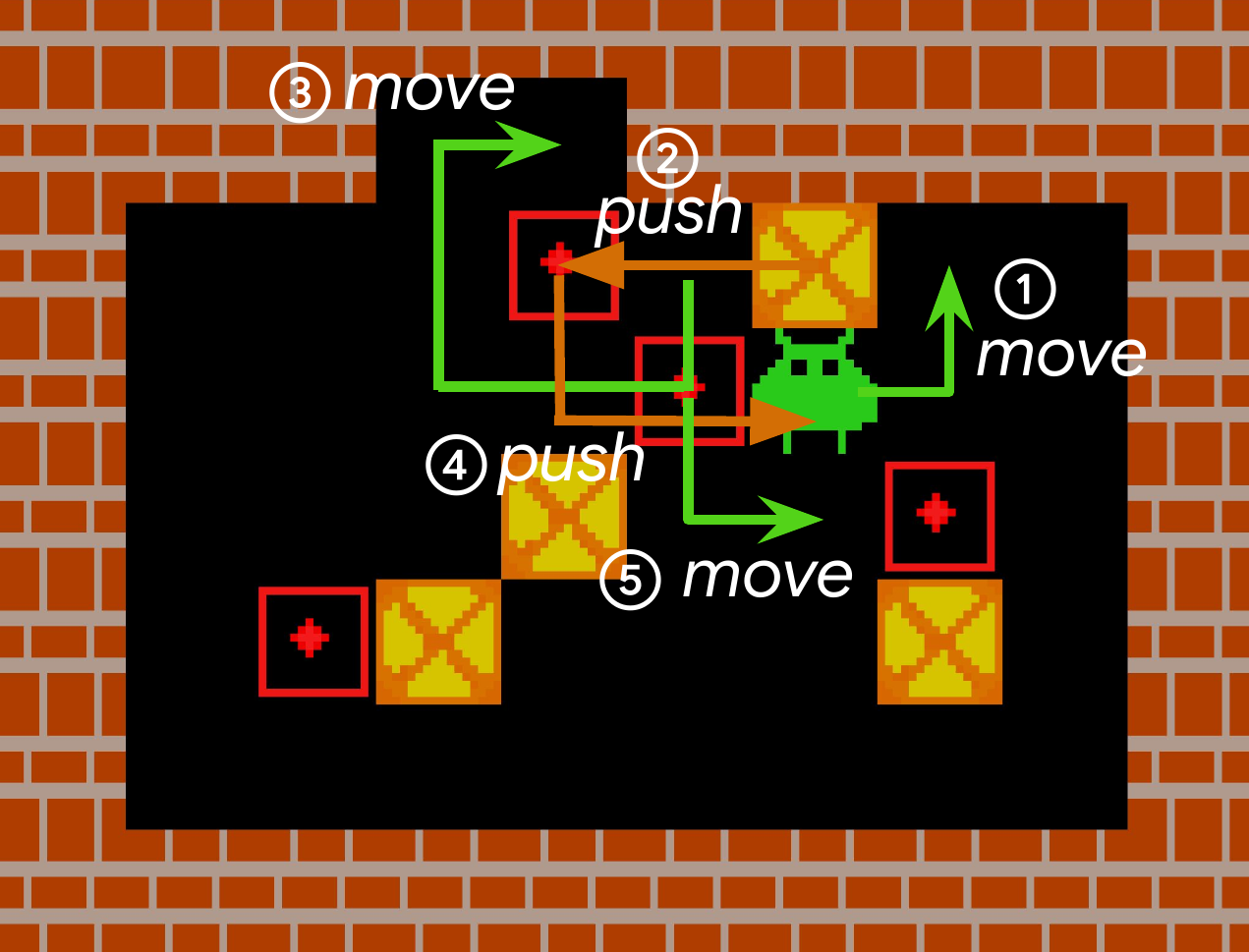}
  \label{fig:sokoban_example}
\end{subfigure}%
\begin{subfigure}{.37\textwidth}
  \centering
  \includegraphics[width=1\linewidth]{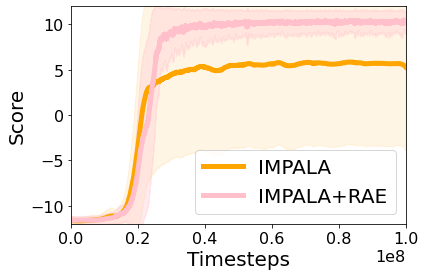}
  \label{fig:sokoban_scores}
\end{subfigure}%
\begin{subfigure}{.37\textwidth}
  \centering
  \includegraphics[width=1\linewidth]{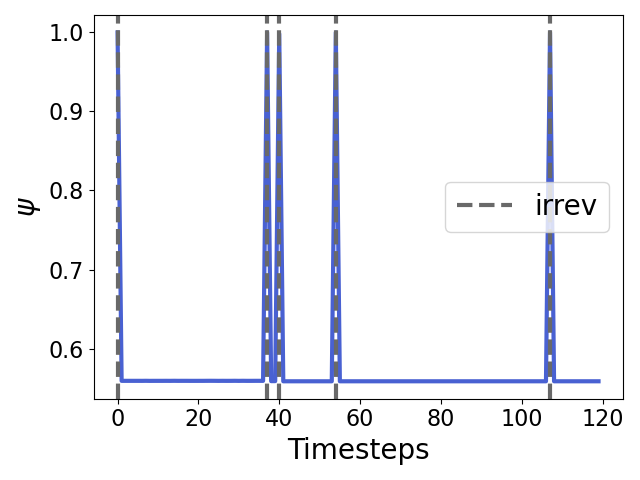}
  \label{fig:sokoban_episode}
\end{subfigure}%
\caption{\textbf{(a):} Non-trivial reversibility: pushing the box against the wall can be reversed by pushing it to the left, going around, pushing it down and going back to start. A minimum of 17 moves is required to go back to the starting state. \textbf{(b):} Performances of IMPALA and IMPALA+RAE on 1k levels of Sokoban (5 seeds average). \textbf{(c):} Evolution of the estimated reversibility along one episode.}
\label{fig:sokoban}
\end{figure}


We benchmark performance on a set of 1k levels. Results are displayed in Fig.~\ref{fig:sokoban}. Equipping an IMPALA agent~\citep{espeholt2018} with RAE leads to a visible performance increase, and the resulting agent consistently solves all levels from the set. We take a closer look at the reversibility estimates and show that they match the ground truth with high accuracy, despite the high imbalance of the distribution (\textit{i.e.} few irreversible transitions, see Fig.~\ref{fig:sokoban}c) and complex reversibility estim ation (see Fig.~\ref{fig:sokoban}a).

\subsection{Safe Control}

In this section, we put an emphasis on RAC, which is particularly suited for safety related tasks.

\begin{figure}
\centering
\begin{subfigure}{.33\textwidth}
  \centering
  \includegraphics[width=1\linewidth]{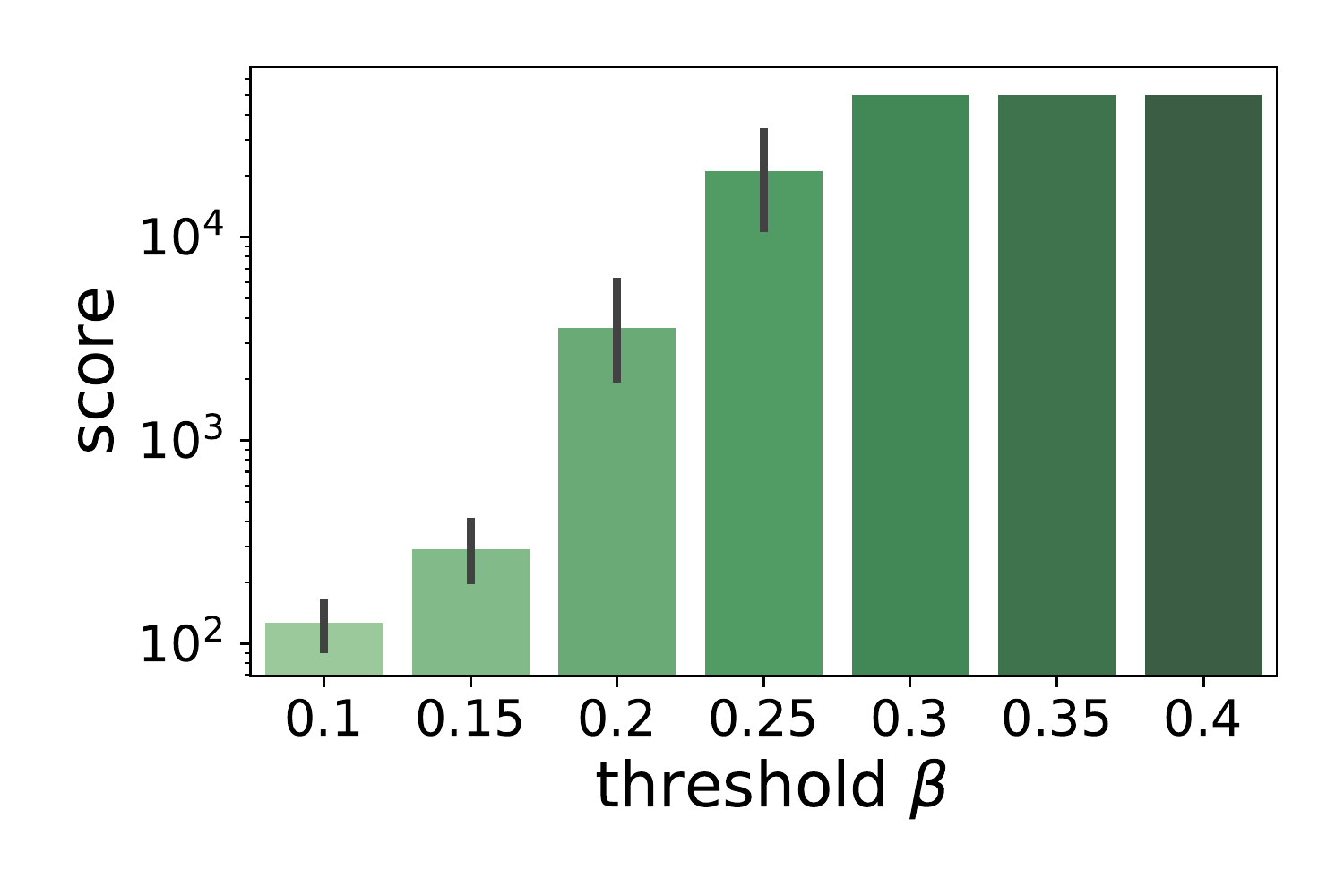}
  \caption{Trajectory lengths}
  \label{fig:cartpole_safe_length}
\end{subfigure}%
\begin{subfigure}{.33\textwidth}
  \centering
  \includegraphics[width=1\linewidth]{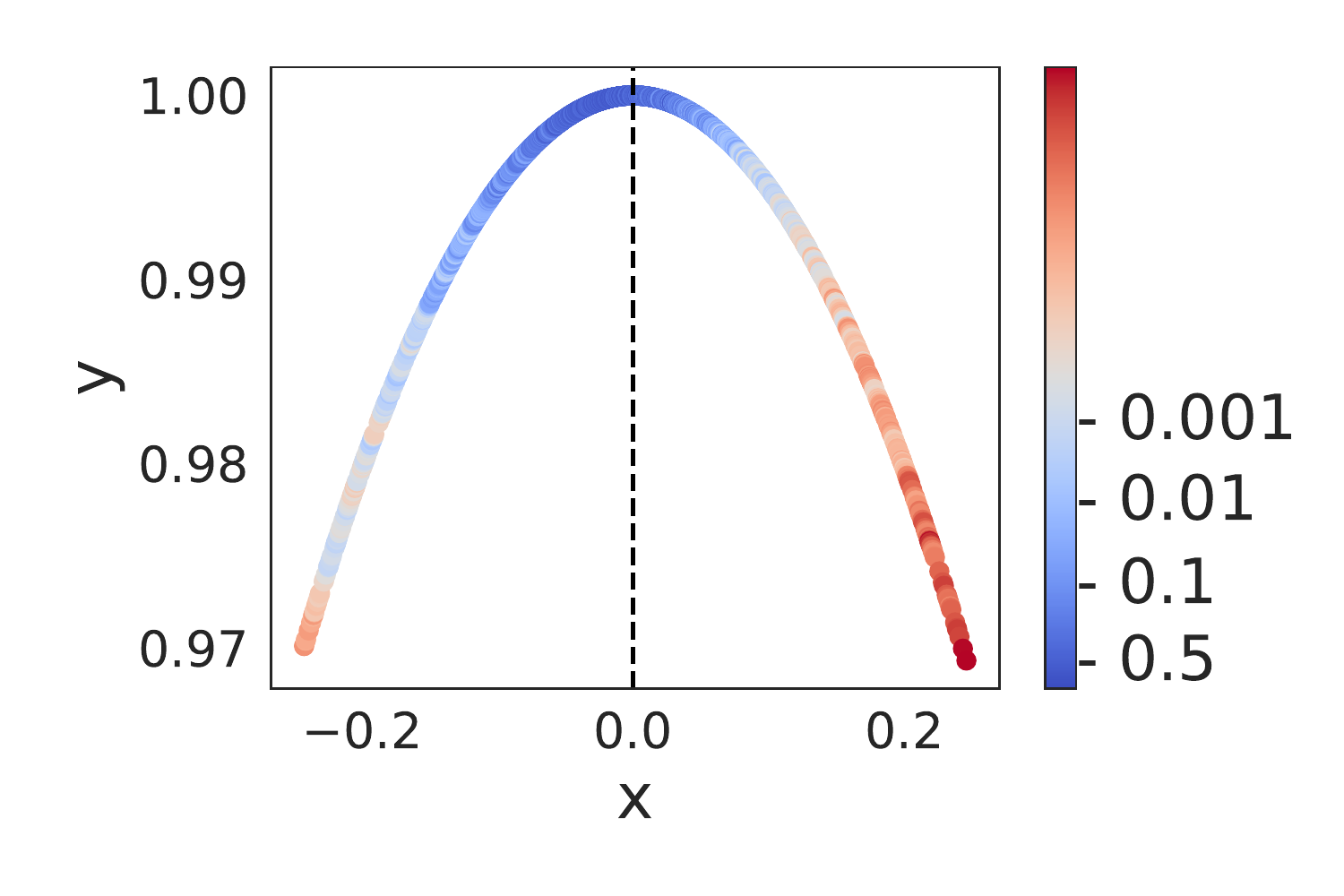}
  \caption{Coordinates: action 0}
  \label{fig:cartpole_a0}
\end{subfigure}%
\begin{subfigure}{.33\textwidth}
  \centering
  \includegraphics[width=1\linewidth]{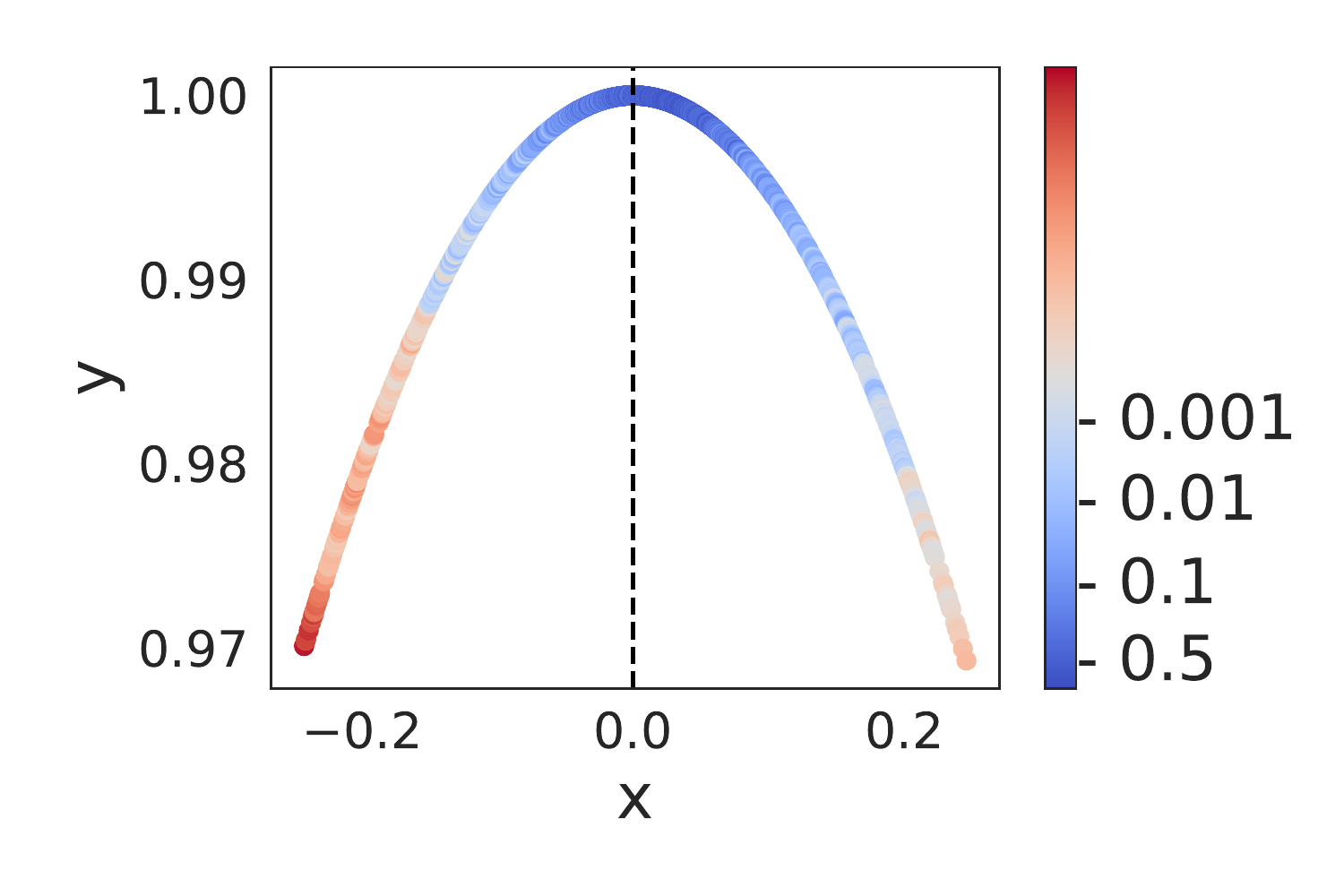}
  \caption{Coordinates: action 1}
  \label{fig:cartpole_a1}
\end{subfigure}
\caption{\textbf{(a):} Mean score of a random policy augmented with RAC on Cartpole+ for several threshold values, with 95\% confidence intervals over 10 random seeds (log scale). \textbf{(b) and (c):}  The $x$ and $y$ axes are the coordinates of the end of the pole relatively to the cart position. The color indicates the estimated reversibility values.}
\label{fig:cartpole_safe}
\end{figure}

\paragraph{Cartpole+.}
We use the standard Cartpole environment, except that we change the maximum number of steps from 200 to 50k to study long-term policy stability. We name this new environment Cartpole+. It is substantially more difficult than the initial setting. We learn reversibility offline, using trajectories collected from a random policy. Fig.~\ref{fig:cartpole_safe_length} shows that a random policy augmented with RAC achieves seemingly infinite scores. For the sake of comparison, we indicate that a DQN~\citep{mnih2015humanlevel} and the state-of-the-art M-DQN~\citep{Vieillard2020} achieve a maximum score of respectively 1152 and 2801 under a standard training procedure, described in Appendix~\ref{appendix:cartpoleDQN}. This can be surprising, since RAC was only trained on random thus short trajectories (mean length of 20). We illustrate the predictions of our learned estimator in Fig.~\ref{fig:cartpole_a0},\ref{fig:cartpole_a1}. When the pole leans to the left ($x$ < 0), we can see that moving the cart to the left is perceived as more reversible than moving it to the right. This is key to the good performance of RAC and perfectly on par with our understanding of physics: when the pole is leaning in a direction, agents must move the cart in the same direction to stabilize it.

\paragraph{Turf.}

\begin{wrapfigure}[15]{r}{0.35\textwidth}
  \begin{center}
    \vspace{-20pt}
    \includegraphics[width=\linewidth]{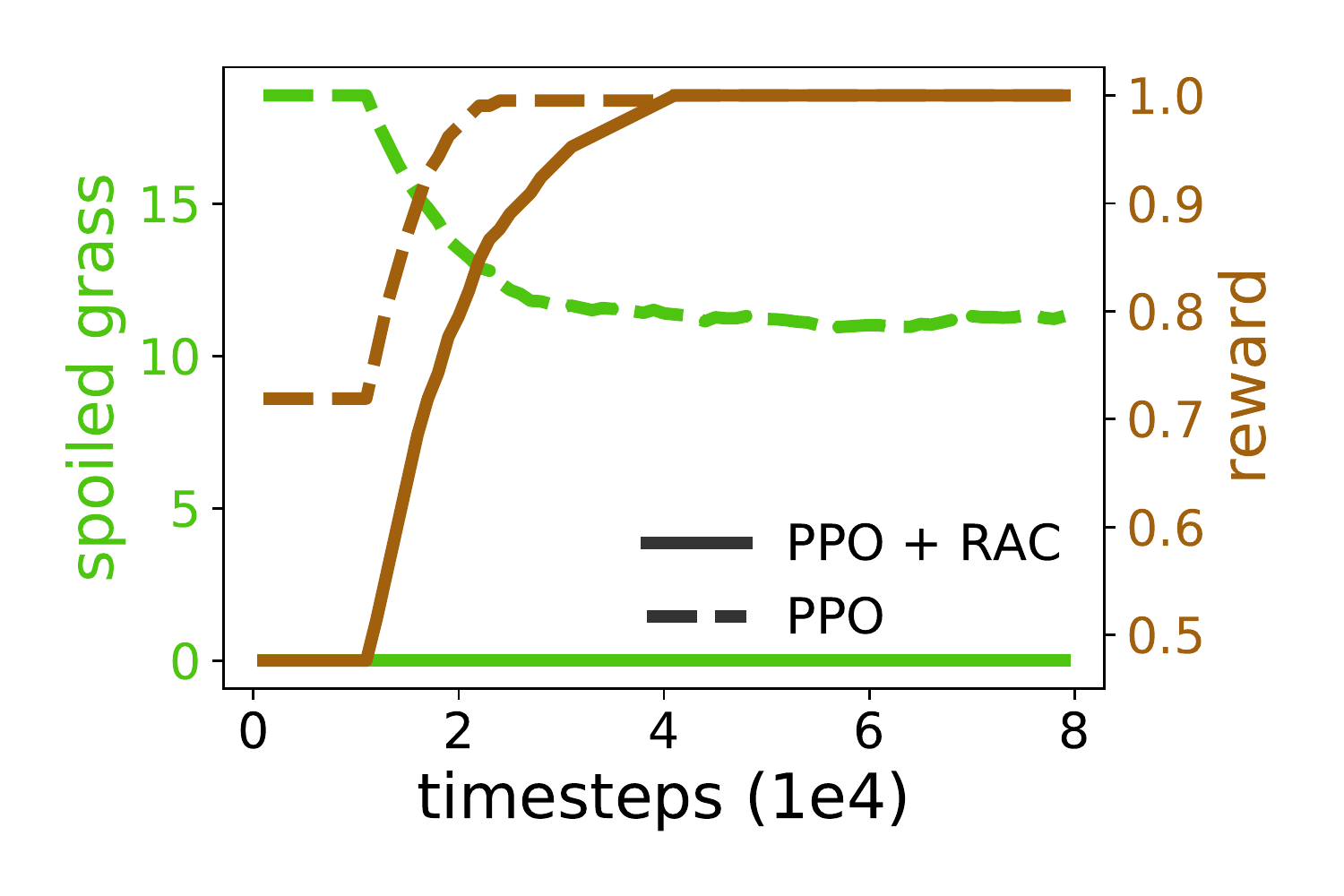}
  \end{center}
  \caption{\looseness=-1 PPO and RAC (solid lines) vs PPO (dashed lines). At the cost of slower learning (brown), our approach prevents the agent from producing a single irreversible side-effect (green) during the learning phase. Curves are averaged over 10 runs.}
  \label{fig:grassland_safe}
\end{wrapfigure}

\looseness=-1
We now illustrate how RAC can be used for safe online learning: the implicitly safe constraints provided by RAC prevent policies from deviating from safe trajectories. This ensures for example that agents stay in recoverable zones during exploration.

We learn the reversibility estimator offline, using the trajectories of a random policy. We reject actions whose reversibility is deemed inferior to $\beta = 0.2$, and train a PPO agent with RAC. As displayed in Fig.~\ref{fig:grassland_safe}, PPO with RAC learns to reach the goal without causing any irreversible side-effect (\textit{i.e.} stepping on grass) during the whole training process.

The threshold $\beta$ is a very important parameter of the algorithm. Too low a threshold could lead to overlooking some irreversible actions, while a high threshold could prevent the agent from learning the new task at hand. We discuss this performance/safety trade-off in more details in Appendix.~\ref{appendix:grassland_tradeoff}.

\section{Conclusion}

In this work, we formalized the link between the reversibility of transitions and their temporal order, which inspired a self-supervised procedure to learn the reversibility of actions from experience. In combination with two novel reversibility-aware exploration strategies, RAE for directed exploration and RAC for directed control, we showed the empirical benefits of our approach in various scenarios, ranging from safe RL to risk-averse exploration. Notably, we demonstrated increased performance in procedurally-generated Sokoban puzzles, which we tied to more efficient exploration.   

\paragraph{Broader impact and ethical considerations.}
\label{subsection:broader-impact}

The presented work aims at estimating and controlling potentially irreversible behaviors in RL agents. We think it has interesting applications in safety-first scenarios, where irreversible behavior or side-effects are to be avoided. The societal implication of these effects would be safer interactions with RL-powered components (\textit{e.g.} robots, virtual assistants, recommender systems) which, though rare today, could become the norm.
We argue that further research and applications should verify that the induced reversible behavior holds in almost all situations and does not lead to unintended effects.
Our method could be deflected from its goal and used to identify and encourage actions with irreversible effects. In this case, a counter measure consists in using our method to flag and replace irreversible actions. Hence, while the method provides information that could be used to deal irreversible harm, we argue that it provides equal capabilities for detection and prevention.

\begin{ack}
Experiments presented in this paper were partially carried out using the Grid’5000 testbed, supported
by a scientific interest group hosted by Inria and including CNRS, RENATER and several Universities
as well as other organizations. NG is a recipient of PhD funding from the AMX program, Ecole
Polytechnique. The authors would like to thank Edouard Leurent, Antoine Moulin, Odalric-Ambrym Maillard, L\'eonard Hussenot, Nino Vieillard, Alexis Jacq, Th\'eophane Weber and Bobak Shahriari for helpful comments and suggestions.
\end{ack}

\newpage
\bibliography{biblio}
\bibliographystyle{abbrvnat}

\newpage
\section*{Checklist}

\begin{enumerate}

\item For all authors...
\begin{enumerate}
  \item Do the main claims made in the abstract and introduction accurately reflect the paper's contributions and scope?
    \answerYes{}
  \item Did you describe the limitations of your work?
    \answerYes{Practical caveats and limitations are discussed at the beginning of Sec.~\ref{par:limitations} and in Sec.~\ref{subsection:broader-impact}.}
  \item Did you discuss any potential negative societal impacts of your work?
    \answerYes{See Sec.~\ref{subsection:broader-impact}.}
  \item Have you read the ethics review guidelines and ensured that your paper conforms to them?
    \answerYes{}
\end{enumerate}

\item If you are including theoretical results...
\begin{enumerate}
  \item Did you state the full set of assumptions of all theoretical results?
    \answerYes{Theoretical assumptions are described along the propositions and theorems in Sec.~\ref{sec:ReversibilityEstimation_via_Classification}.}
	\item Did you include complete proofs of all theoretical results?
    \answerYes{Complete proofs are given in Appendix~\ref{appendix:maths}.}
\end{enumerate}

\item If you ran experiments...
\begin{enumerate}
  \item Did you include the code, data, and instructions needed to reproduce the main experimental results (either in the supplemental material or as a URL)?
    \answerYes{We released the code for every experiments except Sokoban (Sec.~\ref{subsection:sokoban}) as some components are proprietary. Additionally, the pseudo-code for the proposed methods is given in Appendix~\ref{appendix:extra_details_rae_rac}. }
  \item Did you specify all the training details (e.g., data splits, hyperparameters, how they were chosen)?
    \answerYes{See Appendix~\ref{appendix:experiment_details}.}
	\item Did you report error bars (e.g., with respect to the random seed after running experiments multiple times)?
    \answerYes{}
	\item Did you include the total amount of compute and the type of resources used (e.g., type of GPUs, internal cluster, or cloud provider)?
    \answerNo{}
\end{enumerate}

\item If you are using existing assets (e.g., code, data, models) or curating/releasing new assets...
\begin{enumerate}
  \item If your work uses existing assets, did you cite the creators?
    \answerYes{}
  \item Did you mention the license of the assets?
    \answerNo{}
  \item Did you include any new assets either in the supplemental material or as a URL?
    \answerYes{}
  \item Did you discuss whether and how consent was obtained from people whose data you're using/curating?
    \answerNA{}
  \item Did you discuss whether the data you are using/curating contains personally identifiable information or offensive content?
    \answerNA{}
\end{enumerate}

\item If you used crowdsourcing or conducted research with human subjects...
\begin{enumerate}
  \item Did you include the full text of instructions given to participants and screenshots, if applicable?
    \answerNA{}
  \item Did you describe any potential participant risks, with links to Institutional Review Board (IRB) approvals, if applicable?
    \answerNA{}
  \item Did you include the estimated hourly wage paid to participants and the total amount spent on participant compensation?
    \answerNA{}
\end{enumerate}

\end{enumerate}


\newpage

\appendix

\looseness=-1
We organize the supplementary material as follows: in Appendix~\ref{appendix:maths}, we include the proofs of results from the main text, as well as additional formalism; in Appendix~\ref{appendix:extra_details_rae_rac}, we provide additional details about the proposed algorithms, including pseudo-code and figures that did not fit in the main text; and in Appendix~\ref{appendix:experiment_details}, we detail our experimental procedure, including hyperparameters for all methods/

\section{Mathematical Elements and Proofs}
\label{appendix:maths}


\subsection{Possible Definitions of Reversibility}
\label{appendix:definitions_reversibility}

In this section, we present several intuitive definitions of reversibility in MDPs. We chose the third definition as our reference, which we argue presents several advantages over the others, although they can be interesting in specific contexts. Indeed, Eq.~\eqref{eq:undiscounted_reward} is simpler than Eq.~\eqref{eq:discounted_reward}, as it does not depend on the discount factor, and more general than Eq.~\eqref{eq:fixed_ts}, as it does not enforce a fixed number of timesteps for going back to the starting state.

\paragraph{Discounted Reward.}
\begin{align}
\label{eq:discounted_reward}
\phi_{\pi, K}(s, a) & \coloneqq  \sum^{K}_{k > t} \gamma^{k-t} p_\pi(s_{t+k} = s \mid s_t = s, a_t=a)\,, \\
\phi_\pi(s, a) & \coloneqq  \sum^{\infty}_{k > t} \gamma^{k-t}  p_\pi( s_{t+k} = s \mid s_t = s, a_t=a).
\end{align}

\paragraph{Fixed Time Step.}
\begin{align}
\label{eq:fixed_ts}
\phi_{\pi, K}(s, a) & \coloneqq  \sup_{k \leq K} p_\pi(s_{t+k} = s \mid s_t = s, a_t=a)\,, \\
\phi_\pi(s, a) & \coloneqq  \sup_{k \in \mathbb{N}} p_\pi(s_{t+k} = s \mid s_t = s, a_t=a).
\end{align}

\paragraph{Undiscounted Reward.}
\begin{align}
\label{eq:undiscounted_reward}
\phi_{\pi, K}(s, a) &\coloneqq  \sum^{K}_{k = 1} p_\pi(s_{t+k} = s, s_{t+k-1} \ne s , \dots,s_{t+1} \ne s  \mid s_t = s, a_t=a) \nonumber\,, \\ 
                 &= p_\pi(s \in \tau_{t+1:t+K+1}\mid s_t = s, a_t=a)\,. \\
\phi_\pi(s, a) &\coloneqq  \sum^{\infty}_{k = 1} p_\pi( s_{t+k} = s, s_{t+k-1} \ne s , \dots,s_{t+1} \ne s  \mid s_t = s, a_t=a) \nonumber\,, \\
               &= p_\pi(s \in \tau_{t+1:\infty}\mid s_t = s, a_t=a).
\end{align}



\subsection{Additional Properties}
\label{appendix:extra_properties}

\begin{wrapfigure}[10]{r}{0.2\textwidth}
  \begin{center}
    \vspace{-55pt}
    \includegraphics[width=\linewidth]{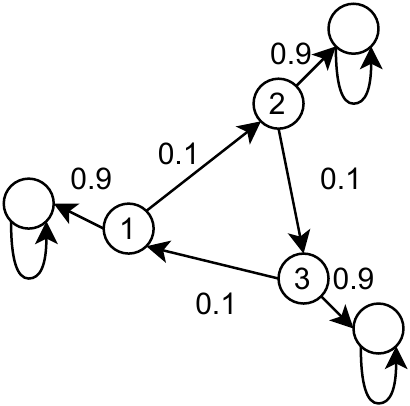}
  \end{center}
  \caption{Counter-example for additional property 4. The initial state is sampled uniformly amongst $\{0, 1, 2\}$.}
  \label{fig:4_couter_example}
\end{wrapfigure}

We write $s \rightarrow s'$ if $\psi_\pi(s, s') \geq 0.5$ ("it is more likely to go from $s$ to $s'$ than to go from $s'$ to $s$") and $s \Rightarrow s'$ if $\psi_\pi(s, s') = 1$ ("it is possible to go from $s$ to $s'$, but it is not possible to come back to $s$ from $s'$").

\begin{enumerate}
    \item $\psi_\pi(s, s') + \psi_\pi(s', s) = 1$
    \item if $s_0 \Rightarrow s_1 \Rightarrow s_2 $ then $s_0 \Rightarrow s_2$ (transitivity for $\Rightarrow$)
    \item if $s_0 \rightarrow s_1 \rightarrow \dots \rightarrow s_i \Rightarrow s_{i+1} \rightarrow \dots \rightarrow s_t$ then $s_0 \Rightarrow s_t$
    \item in general $s_1 \rightarrow s_2$ and $s_2 \rightarrow s_3$ doesn't imply $s_1 \rightarrow s_3$
\end{enumerate}


Proofs:

(1) $\psi_\pi(s, s') + \psi_\pi(s', s) = \mathbb{E}_{\tau \sim \pi} \mathbb{E}_{t\ne t' \vert s_{t}=s, s_{t'}=s'} \big[ \mathbb{1}_{t'>t} + \mathbb{1}_{t'<t} \big] = \mathbb{E}_{\tau \sim \pi} \mathbb{E}_{t\ne t' \vert s_{t}=s, s_{t'}=s'} \big[1 \big] = 1$.

(2) and (3): As (3) is stronger than (2), we only prove (3). If it is possible to have $s_0$ after $s_t$ in a trajectory, then it is possible to have $s_i$ after $s_t$. As we have a positive probability of seeing $s_t$ after $s_{i+1}$, we have a positive probability of seeing $s_i$ after $s_{i+1}$, which contradicts $s_i \Rightarrow s_{i+1}$.

(4) A counter example can be found in Fig.~\ref{fig:4_couter_example}. In this case we clearly have $s_1 \rightarrow s_2$, $s_2 \rightarrow s_3$ and $s_3 \rightarrow s_1$.

\subsection{Proofs of Theorem 1 and Theorem 2}
\label{appendix:proofs_both_theorem}

In the following, we prove simultaneously Theorem~\ref{theorem:convergence} and Theorem~\ref{theorem:inequality_over_2}. We begin by two lemmas. 
\setcounter{lemma}{0}

\begin{lemma}\label{proof:1}
\label{lemma:cardinal}
Given a trajectory $\tau$, we denote by ${\#}_T(s \rightarrow s')$ the number of pairs $(s, s')$ in $\tau_{1:T}$ such that $s$ appears before $s'$. We present a simple formula for $\psi(s', s)$ according to the structure of the state trajectory:
\begin{equation}
     {\psi}_{\pi, T}(s, s') = \frac{\mathbb{E}_{\tau \sim \pi} \big[ {\#}_T(s \rightarrow s') \big] }{\mathbb{E}_{\tau \sim \pi} \big[ {\#}_T(s \rightarrow s') + {\#}_T(s' \rightarrow s) \big] }\,.
\end{equation}
\end{lemma}

\begin{proof}
In order to simplify the notations, we leave implicit the fact that indices are always sampled within $[0, T]$.

\begin{align}
\psi_{\pi, T}(s, s') &= \mathbb{E}_{\pi} \mathbb{E}_{t \ne t' \vert s_{t}=s, s_{t'}=s'} \big[ \mathbb{1}_{t'>t} \big]\,, \\
        &= \frac{\mathbb{E}_{\pi} \mathbb{E}_{t \ne t'} \big[ \mathbb{1}_{t'>t} \mathbb{1}_{s_{t}=s} \mathbb{1}_{s_{t'}=s'} \big]}{\mathbb{E}_{\pi} \mathbb{E}_{t \ne t'} \big[\mathbb{1}_{s_{t}=s} \mathbb{1}_{s_{t'}=s'} \big]}\,.  \\
\end{align}

Similarly, we have:
\begin{equation}
    \mathbb{E}_{\pi} \mathbb{E}_{t'>t} \big[\mathbb{1}_{s_{t}=s} \mathbb{1}_{s_{t'}=s'} \big] = \frac
    {\mathbb{E}_{\pi} \mathbb{E}_{t \ne t'} \big[\mathbb{1}_{t'>t} \mathbb{1}_{s_{t}=s} \mathbb{1}_{s_{t'}=s'} \big]}{\mathbb{E}_{t \ne t'}\big[\mathbb{1}_{t'>t} \big]}\,.
\end{equation}

Combining it with our previous equation:

\begin{align}
        \psi_{\pi, T}(s, s') &= \frac{\mathbb{E}_{\pi} \mathbb{E}_{t'>t} \big[\mathbb{1}_{s_{t}=s} \mathbb{1}_{s_{t'}=s'}\big] \mathbb{E}_{t \ne t'}\big[\mathbb{1}_{t'>t} \big]}{\mathbb{E}_{\pi} \mathbb{E}_{t \ne t'}, \big[\mathbb{1}_{s_{t}=s} \mathbb{1}_{s_{t'}=s'} \big]}\,,  \\
        &= \frac{1}{2} \frac{\mathbb{E}_{\pi} \mathbb{E}_{t'>t} \big[\mathbb{1}_{s_{t}=s} \mathbb{1}_{s_{t'}=s'} \big]}{\mathbb{E}_{\pi} \mathbb{E}_{t \ne t'} \big[\mathbb{1}_{s_{t}=s} \mathbb{1}_{s_{t'}=s'} \big]}\,.  \\
\end{align}

Looking at the denominator, we can notice:

\begin{align}
    {\mathbb{E}_{\pi} \mathbb{E}_{t \ne t'} \big[\mathbb{1}_{s_{t}=s} \mathbb{1}_{s_{t'}=s'} \big]} 
    &=
    \frac{1}{2} {\mathbb{E}_{\pi} \mathbb{E}_{t < t'} \big[\mathbb{1}_{s_{t}=s} \mathbb{1}_{s_{t'}=s'} \big]} +
    \frac{1}{2} {\mathbb{E}_{\pi} \mathbb{E}_{t' < t} \big[\mathbb{1}_{s_{t}=s} \mathbb{1}_{s_{t'}=s'} \big]}\,,
    \\
    &=
    \frac{1}{2} {\mathbb{E}_{\pi} \mathbb{E}_{t < t'} \big[\mathbb{1}_{s_{t}=s} \mathbb{1}_{s_{t'}=s'} + \mathbb{1}_{s_{t}=s'} \mathbb{1}_{s_{t'}=s} \big]}\,,
\end{align}

which comes from the fact that $t$ and $t'$ play a symmetrical role. Thus,

\begin{equation}
        \psi_{\pi, T}(s, s') =
        \frac{
        \mathbb{E}_{\tau \sim \pi} \mathbb{E}_{t} \mathbb{E}_{t' > t} \big[\mathbb{1}_{s_t=s}\mathbb{1}_{s_{t'}=s'} \big]
        }
        {
        \mathbb{E}_{\tau \sim \pi} \mathbb{E}_{t} \mathbb{E}_{t' > t} \big[\mathbb{1}_{s_t=s}\mathbb{1}_{s_{t'}=s'} + \mathbb{1}_{s_t=s'}\mathbb{1}_{s_{t'}=s}\big]
        }\,.
\end{equation}

Since 
\begin{align}
     \mathbb{E}_{\tau \sim \pi} \big[ {\#}_T(s \rightarrow s') \big]
     &=
     \sum_{i < j \leq T} \mathbb{1}_{s_{i}=s} \mathbb{1}_{s_{j}=s'}\,,
     \\
     &= 
     \binom{T}{2} \sum_{i < j \leq T} \frac{1}{\binom{T}{2}} \mathbb{1}_{s_{i}=s} \mathbb{1}_{s_{j}=s'}\,,
     \\
     &= \binom{T}{2} \mathbb{E}_{\tau \sim \pi} \mathbb{E}_{t} \mathbb{E}_{t' > t} \big[\mathbb{1}_{s_t=s}\mathbb{1}_{s_{t'}=s'}\big]\,,
\end{align}

we get: 

\begin{align}
    {\psi}_{\pi, T}(s, s') &= \frac{\binom{T}{2} \mathbb{E}_{\tau \sim \pi} \big[ {\#}_T(s \rightarrow s') \big]}{\binom{T}{2} \mathbb{E}_{\tau \sim \pi} \big[{\#}_T(s \rightarrow s') + {\#}_T(s' \rightarrow s)\big] }\,,
    \\
   {\psi}_{\pi, T}(s, s') &= \frac{\mathbb{E}_{\tau \sim \pi} \big[ {\#}_T(s \rightarrow s') \big]}{\mathbb{E}_{\tau \sim \pi} \big[{\#}_T(s \rightarrow s') + {\#}_T(s' \rightarrow s)\big] }\,.
   \\
\end{align}
\end{proof}

\begin{lemma}
\label{lemma:trajectory}
\looseness=-1 Assume that we are given a fixed trajectory where $s$ appears $k \in \mathbb{N}$ times, in the form of :
\begin{equation}
    s_0 \underbrace{\longrightarrow}_{n_0(s')} s \underbrace{\longrightarrow}_{n_1(s')} s \underbrace{\longrightarrow}_{n_2(s')} s \underbrace{\longrightarrow}_{n_3(s')} \dots \underbrace{\longrightarrow}_{n_{k-1}(s')} s \underbrace{\longrightarrow}_{n_k(s')} \,,
\end{equation}
where $n_i(s')$ denotes the number of times $s'$ appears between the $i^\text{th}$ and the $(i+1)^\text{th}$ occurrence of $s$.
In this case, 
\begin{equation}
    \label{Equ_cardinal}
    {\#}_T(s \rightarrow s') = \sum^k_{i=0} i \times n_i(s')\,.
\end{equation}
If we suppose that $n_1(s') = n_2(s') = \dots = n_{k-1}(s')$, we also have
\begin{equation}
    \label{Equ_doublecardinal}
    {\#}_T(s \rightarrow s') - {\#}_T(s' \rightarrow s) = k \, \big( n_k(s') - n_0(s') \big)\,.
\end{equation}
\end{lemma}

\begin{proof}
Eq.~\eqref{Equ_cardinal} comes directly from 
${\#}_T(s \rightarrow s') = \sum_{i=1}^k \sum_{j=i}^k n_j(s') = \sum^k_{i=0} i \times n_i(s')$.
To prove Equ. \eqref{Equ_doublecardinal}, we first notice that $ {\#}_T(s \rightarrow s') + {\#}_T(s' \rightarrow s) = k \times \sum_{i=0}^k n_i(s') $. Thus 
\begin{align}
    {\#}_T(s \rightarrow s') - {\#}_T(s' \rightarrow s) &= 2 \times {\#}_T(s \rightarrow s') - \big({\#}_T(s \rightarrow s') + {\#}_T(s' \rightarrow s) \big)\,,  \\
    &= 2 \left(k \, n_k(s') + n_1(s') \sum_{i=0}^{k-1} i\right) - \big( k \, n_k(s') + k \, n_0(s') + k \, (k-1) \, n_1(s') \big),\\
    &= k \, n_k(s') - k \, n_0(s')\,.
\end{align}
\end{proof}

\setcounter{theorem}{0}
\begin{theorem}
For every policy $\pi$ and $s, s' \in S$, ${\psi}_{\pi, T}(s, s')$ converges when $T$ goes to infinity.
\end{theorem}

\begin{theorem}
Given a policy $\pi$, a state $s$, and an action $a$, we can link reversibility and empirical reversibility with the inequality: $\Bar{\phi}_\pi(s, a) \geq \frac{\phi_\pi(s, a)}{2} $.
\end{theorem}

\begin{proof}
For a policy $\pi$ and $s, s' \in S$, we define $\Hat{\phi}_\pi(s, s')$ the quantity $p_\pi(s \in \tau_{t+1:\infty}\mid s_t = s')$ such that
$\phi_\pi(s, a) = \mathbb{E}_{s' \sim P(s, a)}\left[\Hat{\phi}_\pi(s, s')\right]$
In order to prove the theorem, we first prove that ${\psi}_T(s', s)$ converges to a quantity denoted by $\psi(s', s)$, and that: 

\begin{equation}
\label{inequ_subcase}
    \forall s, s' \in S, \frac{\Hat{\phi}^\pi(s, s')}{2} \leq \psi(s', s)\,.
\end{equation}
\looseness=-1
We subdivide our problem into four cases, depending on whether $s$ and $s'$ are recurrent or transient. 

\paragraph{Case 1: }
$p_\pi(s \in \tau_{t+1:\infty} \mid s_t = s) < 1$ and $p_\pi(s' \in \tau_{t+1:\infty} \mid s_t = s') = 1$ ($s$ is transient and $s'$ is recurrent for the Markov chain induced by $\pi$).
Informally, this means that if a trajectory contains the state $s'$ we tend to see $s'$ an infinite number of times, and we only see $s$ a finite number of times in a given trajectory.

This implies $\Hat{\phi}_\pi(s, s') = p_\pi(s \in \tau_{t+1:\infty} \mid s_t = s') = 0$, as recurrent states can only be linked to other recurrent states \citep{norris1998}. It is not possible to find trajectories where $s$ appears after $s'$, thus ${\psi}_T(s', s)=0=\psi(s', s)$. Equ.~\eqref{inequ_subcase} becomes "$0 \leq 0$".

\paragraph{Case 2: }
$p_\pi(s \in \tau_{t+1:\infty} \mid s_t = s) = 1$ and $p_\pi(s' \in \tau_{t+1:\infty} \mid s_t = s') < 1$ ($s$ is recurrent and $s'$ is transient for the Markov chain induced by $\pi$).

As before, this implies $\Hat{\phi}_\pi(s', s) = p_\pi(s' \in \tau_{t+1:\infty} \mid s_t = s) = 0$, and thus it is not possible to see in a trajectory $s$ after $s'$. It implies ${\psi}_T(s', s)=1=\psi(s', s)$, so Equ.~\eqref{inequ_subcase} is verified.

\paragraph{Case 3: }
$p_\pi(s \in \tau_{t+1:\infty} \mid s_t = s) = 1$ and $p_\pi(s' \in \tau_{t+1:\infty} \mid s_t = s') = 1$ ($s$ is recurrent and $s'$ is recurrent for the Markov chain induced by $\pi$).
We denote by $T_k$ the random variable corresponding to the time of the {$k^\text{th}$} visit to $s$.
A trajectory can be represented as follows:
\begin{equation}
    s_0 \underbrace{\longrightarrow}_{n_1(s')} s \underbrace{\longrightarrow}_{n_2(s')} s \underbrace{\longrightarrow}_{n_3(s')} s \underbrace{\longrightarrow}_{n_4(s')} \dots \underbrace{\longrightarrow}_{n_{k}(s')} s = s_{T_k} \underbrace{\longrightarrow}_{n_{k+1}(s')} \,,
\end{equation}

where, writing $\sim$ the equality in distribution, $n_2(s') \sim n_3(s') \sim \dots \sim n_k(s')$ and $\mathbb{E}_\tau n_2(s') = \mathbb{E}_\tau n_3(s') = \dots = \mathbb{E}_\tau n_k(s')$ using the strong Markov property. From Lemma~\ref{lemma:cardinal} we get:

\begin{align}
\label{equ:psi_as12}
        {\psi}_{\pi, T}(s, s') &= \frac{\mathbb{E}_{\tau \sim \pi} \big[ {\#}_T(s \rightarrow s') \big]}{\mathbb{E}_{\tau \sim \pi} \big[ {\#}_T(s \rightarrow s') + {\#}_T(s' \rightarrow s) \big] }\,,\\
        &= \frac{1}{2} \frac{\mathbb{E}_{\tau \sim \pi} \big[ {\#}_T(s \rightarrow s') + {\#}_T(s' \rightarrow s) + {\#}_T(s \rightarrow s') - {\#}_T(s' \rightarrow s) \big ]}{\mathbb{E}_{\tau \sim \pi} \big[ {\#}_T(s \rightarrow s') + {\#}_T(s' \rightarrow s) \big]}\,, \\
        &= \frac{1}{2} + \frac{\mathbb{E}_{\tau \sim \pi} \big[ {\#}_T(s \rightarrow s') - {\#}_T(s' \rightarrow s) \big]}{2 \, \mathbb{E}_{\tau \sim \pi} \big[{\#}_T(s \rightarrow s') + {\#}_T(s' \rightarrow s) \big]}\,.
\end{align}

We can see from Lemma~\ref{lemma:trajectory} :

\begin{equation}
   \mathbb{E}_\tau \big[ {\#}_{T_k}(s \rightarrow s') -{\#}_{T_k}(s' \rightarrow s) \big] = - k \, \mathbb{E}_\tau n_1(s') \,.
\end{equation}

Thus,

\begin{align}
\label{equ:limit}
    \frac{\mathbb{E}_\tau \big[ {\#}_{T_k}(s \rightarrow s') -{\#}_{T_k}(s' \rightarrow s) \big]}{\mathbb{E}_{\tau \sim \pi} \big[ {\#}_{T_k}(s \rightarrow s') + {\#}_{T_k}(s' \rightarrow s) \big]} &= \frac{- k \, \mathbb{E}_\tau n_1(s')}{k \, \mathbb{E}_\tau n_1(s') + k^2 \, \mathbb{E}_\tau n_2(s')} \\
    &\xrightarrow[k \to \infty]{} 0.
\end{align}

Given $t \in \mathbb{N}$ and a trajectory $\tau$, we denote ${\#}_T(s)$ the random variable corresponding to the number of times when $s$ appear before $t$, such that a trajectory has the following structure :

\begin{equation}
    s_0 \underbrace{\longrightarrow}_{n_1(s')} s \underbrace{\longrightarrow}_{n_2(s')} s \underbrace{\longrightarrow}_{n_3(s')} s \underbrace{\longrightarrow}_{n_4(s')} \dots \underbrace{\longrightarrow}_{n_{k}(s')} s = s_{T_{{\#}_T(s)}}  \underbrace{\longrightarrow s_t \longrightarrow}_{n_{k+1}(s')} s = s_{T_{{\#}_T(s)+1}} \,.
\end{equation}

\begin{align}
    \frac{\mathbb{E}_\tau \big[ {\#}_T(s \rightarrow s') -{\#}_T(s' \rightarrow s) \big]}{\mathbb{E}_{\tau} \big[ {\#}_T(s \rightarrow s') + {\#}_T(s' \rightarrow s) \big]} 
    &\leq
    \frac{\mathbb{E}_\tau \big[ {\#}_{{\#}_T(s)}(s \rightarrow s') -{\#}_{{\#}_T(s)}(s' \rightarrow s) \big] + \mathbb{E}_\tau {\#}_T(s) n_{k+1}(s')}{\mathbb{E}_{\tau} {\#}_{{\#}_T(s)}(s \rightarrow s') + \mathbb{E}_{\tau} {\#}_{{\#}_T(s)}(s' \rightarrow s)}\,,
    \\
    &\xrightarrow[T \to \infty]{} 0 \text{ as in Equ. \eqref{equ:limit}}.
\end{align}

And, 

\begin{align}
    \frac{\mathbb{E}_\tau \big[ {\#}_T(s \rightarrow s') -{\#}_T(s' \rightarrow s) \big]}{\mathbb{E}_{\tau} \big[ {\#}_T(s \rightarrow s') + {\#}_T(s' \rightarrow s) \big]} 
    &\geq
    \frac{\mathbb{E}_\tau \big[ {\#}_{{\#}_T(s)}(s \rightarrow s') -{\#}_{{\#}_T(s)}(s' \rightarrow s) \big] - \mathbb{E}_\tau \sum_{i=1}^{{\#}_T(s)+1} n_i(s')}{\mathbb{E}_{\tau}\big[ {\#}_{{\#}_T(s)}(s \rightarrow s') + {\#}_{{\#}_T(s)}(s' \rightarrow s) \big]}\,,
    \\
    &\xrightarrow[T \to \infty]{} 0
\end{align}

Therefore, 

\begin{align}
        \frac{\mathbb{E}_\tau \big[ {\#}_T(s \rightarrow s') -{\#}_T(s' \rightarrow s) \big]}{\mathbb{E}_{\tau} \big[ {\#}_T(s \rightarrow s') + {\#}_T(s' \rightarrow s) \big]}
        &\xrightarrow[T \to \infty]{} 0 \, \text{ , and finally,}
\end{align}

\begin{equation}
        {\psi}_{\pi, T}(s, s') = \frac{1}{2} + \frac{\mathbb{E}_\tau \big[ {\#}_T(s \rightarrow s') -{\#}_T(s' \rightarrow s) \big]}{2 \mathbb{E}_{\tau \sim \pi} \big[ {\#}_T(s \rightarrow s') + {\#}_T(s' \rightarrow s) \big]} \xrightarrow[T \to \infty]{} \frac{1}{2}\,.
\end{equation}

As $\Hat{\phi}_\pi(s, s') = 1$ here, we immediately have $\frac{\Hat{\phi}_\pi(s, s')}{2} = \psi(s', s)$. We can notice that the inequality is tight in this case.

\paragraph{Case 4: }
$p_\pi(s \in \tau_{t+1:\infty} \mid s_t = s) < 1$ and $p_\pi(s' \in \tau_{t+1:\infty} \mid s_t = s') < 1$ ($s$ is transient and $s'$ is transient for the Markov chain induced by $\pi$). To simplify the following formulas, we will write $\alpha = p_\pi(s \in \tau_{t+1:\infty} \mid s_t = s')$. 
Here, we denote by $\#(s)$ the random variable corresponding to the total number of visits of the state $s$, and $\#(s \rightarrow s')$ the number of pairs such that $s$ appears before $s'$. $\#(s)$ follows the geometric distribution $G\left(1 - p_\pi(s \in \tau_{t+1:\infty} \mid s_t = s) \right)$.

${\#}_T(s \rightarrow s')$ converges almost surely to $\#(s \rightarrow s')$, and we have 
${\#}_T(s \rightarrow s') \leq \#(s \rightarrow s')$. Therefore, using the dominated convergence theorem, 
$\mathbb{E}_\tau \big[ {\#}_T(s \rightarrow s') \big] \xrightarrow[T \to \infty]{} \mathbb{E}_\tau \big[ \#(s \rightarrow s')$ \big], and thus:

\begin{equation}
    {\psi}_{\pi, T}(s', s) = \frac{\mathbb{E}_\tau \big[ {\#}_T(s' \rightarrow s) \big]}{ \mathbb{E}_{\tau} \big[ {\#}_T(s \rightarrow s') + {\#}_T(s' \rightarrow s) \big]} \xrightarrow[T \to \infty]{} \frac{\mathbb{E}_\tau {\#}(s' \rightarrow s)}{\mathbb{E}_{\tau} \big[ {\#}(s \rightarrow s') + {\#}(s' \rightarrow s) \big]} = \psi^\pi(s', s) \,.
\end{equation}

This time, we consider a trajectory $\tau$ where $s$ appears $k$ times after $s'$, such that it is of the form:

\begin{equation}
     \bracks{s'}{n_0(s') \geq 0} \longrightarrow \bracks{s}{n_1(s) > 0} \longrightarrow  \bracks{s'}{n_1(s') > 0} \longrightarrow \bracks{s}{n_2(s) >0} \longrightarrow \dots \longrightarrow \bracks{s'}{n_{k-1}(s') >0} \longrightarrow \bracks{s}{n_k(s) >0} \longrightarrow  \bracks{s'}{n_{k}(s') \geq 0} \longrightarrow
\end{equation}

Here, $n_0(s')$ is the number of times when $s'$ appears in the trajectory before the first appearance of $s'$, $n_i(s)$ is the number of times when $s$ appears between two occurrences of $s'$, and $n_{k}(s')$ the number of times when $s'$ appears after the last appearance of $s$. From the strong Markov property, $n_1(s') \sim n_2(s') \sim \dots \sim n_{k-1}(s')$ and $n_1(s) \sim n_2(s) \sim \dots \sim n_{k}(s)$. Note also that these variables are all independent. Here $k$ is a random variable following the geometric distribution $G(\alpha)$ where $\alpha = p(s \in \tau_{t:\infty} \mid s_t=s')$. Notice that when $n_{k}(s') > 0$, we have $n_k(s) \sim n_1(s)$ and $n_k(s') \sim n_1(s')$.



Using these two simplifications, one can write:

\begin{align}
    \mathbb{E}_\tau \Big[ {\#}(s' \rightarrow s) -{\#}(s \rightarrow s') \Big\vert k \Big] &\geq \mathbb{E}_\tau \Big[ {\#}(s' \rightarrow s) - {\#}(s \rightarrow s') \Big\vert k, n_k(s')>0 \Big] \,,
    \\
    &\geq \mathbb{E}_\tau \Big[  n_0(s') \big[ n_1(s) + (k-1) n_1(s) + n_k(s) \big] -
    n_1(s) \big[ k n_1(s') + n_k(s') \big] + \\
    & n_k(s) \big[ k n_1(s') - n_k(s') \big] - 
    n_k(s') (k-1) n_1(s) \Big\vert k, n_k(s')>0 \Big]\,,
    \\ 
    &\geq - k \mathbb{E}_\tau \Big[ n_1(s) \Big\vert k \Big] \mathbb{E}_\tau \Big[ n_k(s') \Big\vert k, n_k(s')>0 \Big] \text{ as in Lemma~\ref{lemma:trajectory}} \,,
    \\
    &\geq -k \mathbb{E}_\tau (n_1(s)) \mathbb{E}_\tau (n_1(s'))\,.
\end{align}

Likewise,

\begin{align}
    \mathbb{E}_\tau \big[ {\#}(s' \rightarrow s) + {\#}(s \rightarrow s') \, \big\vert \, k \big] 
    &= \mathbb{E}_\tau \Big[ k \, n_1(s) n_k(s') + k \, n_0(s') n_1(s) + k \, (k-1) n_1(s) n_1(s') \Big\vert k \Big]\,,\\
    &= k \Big[
    \mathbb{E}_\tau \big[ n_1(s) \big] \mathbb{E}_\tau \big[ n_1(s') \big] + \mathbb{E}_\tau \big[n_1(s) \big] \mathbb{E}_\tau \big[ n_0(s') \big]
    \Big]
    + k (k-1) \mathbb{E}_\tau \big[ n_1(s) \big] \mathbb{E}_\tau \big[ n_1(s') \big]\,.
\end{align}

Thus,

\begin{align}
&\frac{\mathbb{E}_\tau \big[ {\#}(s' \rightarrow s) -{\#}(s \rightarrow s') \big]}{ \mathbb{E}_{\tau} \big[ {\#}(s \rightarrow s') + {\#}(s' \rightarrow s) \big]}
=
\frac{\sum_{i=1}^{\infty} p(k=i) \mathbb{E}_\tau \left[ {\#}(s' \rightarrow s) -{\#}(s \rightarrow s') \, \middle \vert \, k=i \right]}
{\sum_{i=1}^{\infty} p(k=i)  \mathbb{E}_{\tau} \left[ {\#}(s \rightarrow s') + {\#}(s' \rightarrow s) \, \middle \vert \, k=i \right]}\,,
\\&\geq
- \frac{\sum_{i=1}^{\infty} \alpha^{i-1} (1-\alpha) \, i \, \mathbb{E}_\tau \big[n_1(s)\big] \, \mathbb{E}_\tau \big[n_1(s')\big]}
{\sum_{i=1}^{\infty} \alpha^{i-1} (1-\alpha) \Big[ i \left(
    \mathbb{E}_\tau \big[ n_1(s) \big] \mathbb{E}_\tau \big[ n_1(s') \big] + \mathbb{E}_\tau \big[n_1(s)\big] \mathbb{E}_\tau  \big[n_0(s')\big]
    \right)
    + i \, (i-1) \mathbb{E}_\tau \big[ n_1(s) \big] \mathbb{E}_\tau \big[ n_1(s') \big] \Big]}\,,
\\&\geq
- \frac{\sum_{i=1}^{\infty} \alpha^{i-1} (1-\alpha) i \mathbb{E}_\tau \big[n_1(s)\big] \mathbb{E}_\tau \big[n_1(s')\big]}
{\sum_{i=1}^{\infty} \alpha^{i-1} (1-\alpha) \Big[ i \,
    \mathbb{E}_\tau \big[ n_1(s) \big] \mathbb{E}_\tau \big[ n_1(s') \big]
    + i \, (i-1) \mathbb{E}_\tau \big[ n_1(s) \big] \mathbb{E}_\tau \big[ n_1(s') \big] \Big]}\,,
\\&\geq
- \frac{\sum_{i=1}^{\infty} \alpha^{i-1} (1-\alpha) \, i }
{\sum_{i=1}^{\infty} \alpha^{i-1} (1-\alpha) \big[ i
    + i \, (i-1) \big]}\,,
\\&\geq
- \frac{\sum_{i=1}^{\infty} \alpha^{i-1} (1-\alpha) \, i }
{\sum_{i=1}^{\infty} \alpha^{i-1} (1-\alpha) \, i^2 }\,,
\\&\geq - \frac{
\frac{1}{1 -\alpha}
}
{\frac{1+\alpha}{(1-\alpha)^2}
}\,,
\\&\geq - \frac{
1 - \alpha
}
{1 + \alpha
}\,.
\end{align}

From Lemma~\ref{lemma:cardinal}, 

\begin{align}
\psi^\pi(s', s) &= \frac{1}{2} \left(1 + \frac{\mathbb{E}_\tau \big[ {\#}(s' \rightarrow s) -{\#}(s \rightarrow s') \big]}{ \mathbb{E}_{\tau \sim \pi} \big[ {\#}(s \rightarrow s') + {\#}(s' \rightarrow s) \big]} \right)\,,
\\&\geq \frac{1}{2}\left(1 - \frac{1-\alpha}{1+\alpha}\right)\,,
\\&\geq \frac{\alpha}{1 + \alpha}\,,
\\&\geq \frac{\alpha}{2} = \frac{\Hat{\phi}_\pi(s, s')}{2}\,.
\end{align}

As a quick summary, we divided our problem in 4 cases, and proved that in each case, for every pair of states $s, s'$, we have $\psi^\pi(s', s) \geq \frac{\Hat{\phi}_\pi(s, s')}{2}$.

To end the proof, we simply take the expectation over the distribution of the next states:

\begin{align}
    \mathbb{E}_{s' \sim P(s, a)} \psi_\pi(s', s) 
    &\geq 
    \frac{1}{2}
    \mathbb{E}_{s' \sim P(s, a)} \Hat{\phi}_\pi(s, s') \,,
    \\ \Bar{\phi}_\pi(s, a)
    &\geq 
    \frac{\phi_\pi(s,a)}{2}\,.
\end{align}

\end{proof}

\subsection{Proof of proposition 1}
\label{appendix:prop1}

\setcounter{proposition}{0}
\begin{proposition}
We suppose that we are given a state $s$, an action $a$ such that $a$ is reversible in $K$ steps, a policy $\pi$ and $\rho > 0$. Then,  
$\Bar{\phi}_\pi(s, a) \geq \frac{\rho^K}{2} $, where $A$ denotes the number of actions. Moreover, we have for all $K \in \mathbb{N}$:
$\Bar{\phi}_\pi(s, a) \geq \frac{\rho^K}{2} \phi_K(s, a)$.
\end{proposition}

\begin{proof}
We first prove the second part of the proposition, which is more general. From Definition~\ref{definition:degree_of_rev}, and as the set of policies is closed, there is a policy $\pi^*$ such that $\phi_K(s, a) = p_{\pi^*}(s \in \tau_{t+1:t+K+1}\mid s_t = s, a_t=a)$. We begin by noticing that $\pi$ has a probability at least equal to $\rho$ to copy the policy $\pi^*$ in every state.

It can be stated more formally: 
\begin{equation}
     \forall s \in S, \mathbb{E}_{a\sim \pi(s), a^*\sim \pi^*(s)}(\mathbb{1}_{a = a'}) = \sum_{a\in A} p_\pi(a \mid s) p_{\pi^*}(a\mid s) \geq \rho \Big( \sum_{a\in A} p_{\pi^*}(a\mid s) \Big) = \rho\,.
\end{equation}

Then, we have: 

\begin{align}
    &\phi_{\pi, K}(s, a)=
    p_{\pi}(s \in \tau_{t+1:t+K+1}\mid s_t = s, a_t=a) \,,
    \\
    &= \mathbb{E}_{\pi} \big[ \mathbb{1}_{s \in \tau_{t+1:t+K+1}}\mid s_t = s, a_t=a \big] \,,
    \\
    &= \mathbb{E}_{s_{t+2}, \dots, s_{t+K+1} \sim \pi} \mathbb{E}_{s_{t+1} \sim p(s_t, a_t)} \big[\mathbb{1}_{s \in \tau_{t+1:t+K+1}}\mid s_t = s, a_t=a\big] \,,
    \\
   &= \mathbb{E}_{s_{t+3}, \dots, s_{t+K+1} \sim \pi} \mathbb{E}_{\substack{a_{t+1} \sim \pi(s_{t+1})\\ s_{t+2} \sim p(s_{t+1}, a_{t+1})}}
    \mathbb{E}_{s_{t+1} \sim p(s_t, a_t)} \big[\mathbb{1}_{s \in \tau_{t+1:t+K+1}}\mid s_t = s, a_t=a \big] \,,
    \\
    &= \mathbb{E}_{s_{t+3}, \dots, s_{t+K+1} \sim \pi} \mathbb{E}_{\substack{a_{t+1} \sim \pi(s_{t+1}),
    a_{t+1}' \sim \pi^*(s_{t+1})\\ s_{t+2} \sim p(s_{t+1}, a_{t+1})}}
    \mathbb{E}_{s_{t+1} \sim p(s_t, a_t)} \big[\mathbb{1}_{s \in \tau_{t+1:t+K+1}}\mid s_t = s, a_t=a \big] \,,
    \\
    &\geq \mathbb{E}_{s_{t+3}, \dots, s_{t+K+1} \sim \pi} \mathbb{E}_{\substack{a_{t+1} \sim \pi(s_{t+1}),
    a_{t+1}^* \sim \pi^*(s_{t+1})\\ s_{t+2} \sim p(s_{t+1}, a_{t+1})\\
    s_{t+1} \sim p(s_t, a_t)}}
     \big[\mathbb{1}_{s \in \tau_{t+1:t+K+1}}\mid s_t = s, a_t=a, a_{t+1}=a_{t+1}^*\big] \mathbb{1}_{a_{t+1} = a_{t+1}^*} \,,
     \\
    &\geq \rho \, \mathbb{E}_{s_{t+3}, \dots, s_{t+K+1} \sim \pi}
    \mathbb{E}_{s_{t+1}, s_{t+2} \sim \pi^*} \big[\mathbb{1}_{s \in \tau_{t+1:t+K+1}}\mid s_t = s, a_t=a\big] \text{ , and iterating the same process,} \,,
    \\
    &\geq \rho^K
    \mathbb{E}_{s_{t+1}, s_{t+2}, \dots, s_{t+K+1} \sim \pi^*} \big[\mathbb{1}_{s \in \tau_{t+1:t+K+1}}\mid s_t = s, a_t=a\big] \,,
    \\
    &\geq \rho^K \phi_K(s, a) \,.
\end{align}


We can conclude using Theorem~\ref{theorem:inequality_over_2}: 
$\Bar{\phi}_\pi(s, a) \geq \frac{\phi_\pi(s, a)}{2}
\geq \frac{\phi_{\pi, K}(s, a)}{2} \geq 
\frac{\rho^K}{2} \phi_K(s, a) $.

\end{proof}

\section{Additional Details About Reversibility-Aware RL}
\label{appendix:extra_details_rae_rac}

\subsection{Learning a reversibility estimator}

\begin{figure}
  \centering
  \includegraphics[width=\textwidth]{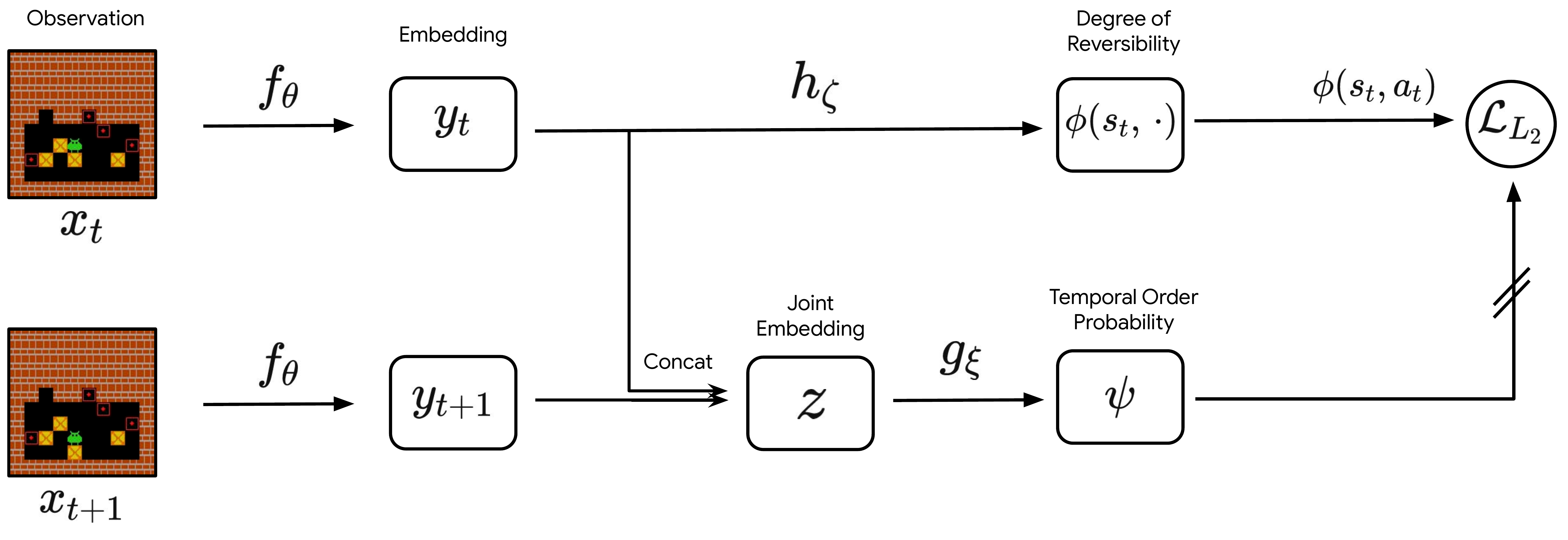}
  \caption{The training procedure for the reversibility estimator used in RAC.}
  \label{fig:action-conditioned-precedence}
\end{figure}

We illustrate how the reversibility estimator is trained in Fig.~\ref{fig:action-conditioned-precedence}. We remind the reader that it is a component that is specific to RAC. See Algorithm~\ref{alg:rac_online} for the detailed procedure of how to train it jointly with the standard precedence estimator and the RL agent.

\subsection{Pseudo-code for RAE and RAC}
\label{appendix:pseudo-code}

\begin{algorithm}
\SetAlgoLined
\caption{RAE: Reversibility-Aware Exploration (online)}
 Initialize the agent weights $\Theta$ and number of RL updates per trajectory $k$\;
 Initialize the precedence classifier weights $\theta, \xi$, window size $w$, threshold $\beta$ and learning rate $\eta$\;
 Initialize the replay buffer $\mathcal{B}$\;
 \For{each iteration}{
  \tcc{Collect interaction data and train the agent.}
  Sample a trajectory $\tau = \{x_i, a_i, r_i\}_{i=1 \ldots T}$ with the current policy\;
  Incorporate irreversibility penalties $\tau' = \big\{x_i, a_i, r_i + r_\beta \big( \psi_{\theta, \xi}(x_i, x_{i + 1}) \big) \big\}_{i=1 \ldots T}$\;
  Store the trajectory in the replay buffer $\mathcal{B} \leftarrow \mathcal{B} \cup \tau$\;
  Do $k$ RL steps and update $\Theta$\; 
  \tcc{Update the precedence classifier.}
  \For{each training step}{
    Sample a minibatch $\mathcal{D}_{batch}$ from $\mathcal{B}$\;
    \tcc{Self-supervised precedence classification, loss in Eq.\eqref{eq:ssl}.}
    $\theta \leftarrow \theta - \eta \nabla_\theta \mathcal{L}_{SSL}(\mathcal{D}_{batch})$\;
    $\xi \leftarrow \xi - \eta \nabla_\xi \mathcal{L}_{SSL}(\mathcal{D}_{batch})$\;
  }
 }
 \label{alg:rae_online}
\end{algorithm}

\begin{algorithm}
\SetAlgoLined
\caption{RAC: Reversibility-Aware Control (online)}
 Initialize the agent weights $\Theta$ and number of RL updates per trajectory $k$\;
 Initialize the precedence classifier weights $\theta, \xi$, window size $w$, threshold $\beta$ and learning rate $\eta$\;
 Initialize the reversibility estimator weights $\zeta$\;
 Initialize the replay buffer $\mathcal{B}$\;
 \For{each iteration}{
  \tcc{Collect interaction data with the modified control policy and train the agent.}
  Sample a trajectory $\tau$ under the rejection sampling policy $\Bar{\pi}$ from eq.\eqref{eq:rejection-sampling} \;
  Store the trajectory in the replay buffer $\mathcal{B} \leftarrow \mathcal{B} \cup \tau$\;
  Do $k$ RL steps and update $\Theta$\; 
  \tcc{Update the precedence classifier.}
  \For{each training step}{
    Sample a minibatch $\mathcal{D}_{batch}$ from $\mathcal{B}$\;
    \tcc{Self-supervised precedence classification, loss in Eq.\eqref{eq:ssl}.}
    $\theta \leftarrow \theta - \eta \nabla_\theta \mathcal{L}_{SSL}(\mathcal{D}_{batch})$\;
    $\xi \leftarrow \xi - \eta \nabla_\xi \mathcal{L}_{SSL}(\mathcal{D}_{batch})$\;
  }
  \tcc{Update the reversibility estimator, loss in Eq.\eqref{eq:regression}.}
  \For{each training step}{
    Sample a minibatch $\mathcal{D}_{batch}$ from $\mathcal{B}$\;
    \tcc{Regression of the precedence classifier probabilities.}
    $\zeta \leftarrow \zeta - \eta \nabla_\zeta \mathcal{L}_{L2}(\mathcal{D}_{batch}, \psi_{\theta, \xi})$\;
  }
 }
 \label{alg:rac_online}
\end{algorithm}

We give the pseudo-code for the online versions of RAE (Algorithm~\ref{alg:rae_online}) and RAC (Algorithm~\ref{alg:rac_online}). The rejection sampling policy $\Bar{\pi}$ under approximate reversibility $\phi$ and threshold $\beta$ is expressed as follows:
\begin{align} \label{eq:rejection-sampling}
    \Bar{\pi}(a | x) =
    \begin{cases}
      0 &\text{ if } \phi(x, a) < \beta \\
      \pi(a | x) / Z &\text{ otherwise, with } Z = \sum_{a' \in \mathcal{A}} \mathbb{1}\{ \phi(x, a') \geq \beta \} \pi(a' | x)
    \end{cases}.
\end{align}
This is equivalent, on average, to sampling from the policy $\pi$ until an action that is reversible enough is found.

The loss we use to train the precedence estimator has the expression:
\begin{equation}
\label{eq:ssl}
    \mathcal{L}_{SSL}(\mathcal{D}_{batch}) = \frac{1}{|\mathcal{D}_{batch}|} \sum_{(x, x', y) \in \mathcal{D}_{batch}} -y \log\big( \psi_{\theta, \xi}(x, x') \big) + (1 - y) \log\big( 1 - \psi_{\theta, \xi}(x, x') \big),
\end{equation}
where $y$ is the binary result of the shuffle, with value $1$ if observations were not shuffled (thus in the correct temporal order), and $0$ otherwise. Pairs of observations $(x, x')$ can be separated by up to $w$ timesteps.

The loss we use to train the reversibility estimator (in RAC only) has the expression:
\begin{equation}
\label{eq:regression}
    \mathcal{L}_{L2}(\mathcal{D}_{batch}, \psi_{\theta, \xi}) = \frac{1}{2 \, |\mathcal{D}_{batch}|} \sum_{(x, a, x') \in \mathcal{D}_{batch}} \big( \psi_{\theta, \xi}(x, x') - \phi_\zeta(x, a) \big)^2,
\end{equation}
where $(x, a, x')$ are triples of state, action and next state sampled from the collected trajectories.

The offline versions of both RAE and RAC can be derived by separating each online algorithm into two parts: 1) training the precedence classifier (and the reversibility estimator for RAC), which is achieved by removing the data collection and RL steps and by using a fixed replay buffer; and 2) training the RL agent, which is the standard RL procedure augmented with modified rewards for RAE, and modified control for RAC, using the classifiers learned in the first part without fine-tuning.

\section{Experimental Details}
\label{appendix:experiment_details}

\subsection{Reward-Free Reinforcement Learning}
\label{appendix:reward_free_rl}

\paragraph{Cartpole.}

The observation space is a tabular 4-dimensional vector: (cart position $x$, cart velocity $ \dot x$, pole angle $\theta$, pole velocity $\dot \theta$). The discrete action space consists of applying a force left or right. The episode terminates if the pole angle is more than ±12° ($\lvert \theta \rvert \leq 0.209$ radians), if the cart position is more than ±2.4, or after 200 time-steps. It is considered solved when the average return is greater than or equal to 195.0 over 100 consecutive trials.

\paragraph{Architecture and hyperparameters.}
The reversibility network inputs a pair of observations and produces an embedding by passing each one into 2 fully connected layers of size 64 followed by ReLU. The two embeddings are concatenated, and projected into a scalar followed by a sigmoid activation.
We trained this network doing 1 gradient step every 500 time steps, using the Adam optimizer~\citep{kingma2014} and a learning rate of 0.01. We used batches of 128 samples, that we collected from a replay buffer of size 1 million. The penalization threshold $\beta$ was fine-tuned over the set $ [0.5, 0.6, 0.7, 0.8, 0.9]$ and eventually set to 0.7. We notice informally that it was an important parameter. A low threshold could lead to over penalizing the agent leading the agent to terminate the episode as soon as possible, whereas a high threshold could slow down the learning. 

Regarding PPO, both the policy network and the value network are composed of two hidden layers of size 64.
Training was done using Adam and a learning rate of 0.01. Other PPO hyperparameters were defaults in \citet{stable-baselines3}, except that we add an entropy cost of 0.05.

\subsection{Learning Reversible Policies}
\label{appendix:learning_reversible_policies}

\paragraph{Environment.}
 The environment consists of a 10 × 10 pixel grid. It contains an agent, represented by a single blue pixel, which can move
in four directions: up, down, left, right. The pink pixel represents the goal, green pixels grass and grey pixels a stone path. Stepping on grass spoils it and the corresponding pixel turns brown, as shown in Fig.~\ref{fig:grassland_traj}.
A level terminates after getting to the goal, or after 120 timesteps. Upon reaching the goal, the agent receives a reward of +1, every other action being associated with 0 reward.

\paragraph{Architecture and hyperparameters.}

The reversibility network takes a pair of observations as input and produces an embedding by passing each observation through 3 identical convolutional layers of kernel size 3, with respectively 32, 64 and 64 channels. The convolutional outputs are flattened, linearly projected onto 64 dimensional vectors and concatenated. The resulting vector is projected into a scalar, which goes through a final sigmoid activation.

As done for Cartpole, we trained this network doing 1 gradient step every 500 time steps, using the Adam optimizer with a learning rate of 0.01. We used minibatches of 128 samples, that we collected from a replay buffer of size 1M. The penalization threshold $\beta$ was set to 0.8, and the intrinsic reward was weighted by 0.1, such that the intrinsic reward was equal to $- 0.1 \, \mathbb{1}_{\psi(s_t, s_{t+1}) >0.8} \, \psi(s_t, s_{t+1})$. 

For PPO, both the policy network and the value network are composed of 3 convolutional layers of size 32, 64, 64. The output is flattened and passed through a hidden layer of size 512. Each layers are followed by a ReLU activation. Policy logits (size 4) and baseline function (size 1) were produced by a linear projection. Other PPO hyperparameters were defaults in~\citet{stable-baselines3}, except that we add an entropy cost of 0.05.

\subsection{Sokoban}

We use the Sokoban implementation from~\citet{schrader2018}.
The environment is a 10x10 grid with a unique layout for each level. The agent receives a -0.1 reward at each timestep, a +1 reward when placing a box on a target, a -1 reward when removing a box from a target, and a +10 reward when completing a level. Observations are of size $(10, 10, 3)$. Episodes have a maximal length of 120, and terminate upon placing the last box on the remaining target.
At the beginning of each episode, a level is sampled uniformly from a set of 1000 levels, which prevents agents from memorizing puzzle solutions. The set is obtained by applying random permutations to the positions of the boxes and the position of the agent, and is pre-computed for efficiency. All levels feature four boxes and targets.

We use the distributed IMPALA implementation from the Acme framework~\citep{hoffman2020} as our baseline agent in these experiments. The architecture and hyperparameters were obtained by optimizing for sample-efficiency on a single held-out level. Specifically, the agent network consists of three 3x3 convolutional layers with 8, 16 and 16 filters and strides 2, 1, and 1 respectively; each followed by a ReLU nonlinearity except the last one. The outputs are flattened and fed to a 2-layer feed-forward network with 64 hidden units and ReLU nonlinearities. The policy and the value network share all previous layers, and each have a separate final one-layer feed-forward network with 64 hidden units and ReLU nonlinearities as well.
Regarding agent hyperparameters, we use 64 actors running in parallel, a batch size of 256, an unroll length of 20, and a maximum gradient norm of 40. The coefficient of the loss on the value is 0.5, and that of the entropic regularization 0.01. We use the Adam optimizer with a learning rate of 0.0005, a momentum decay of 0 and a variance decay of 0.99.

The precedence estimator network is quite similar: it consists of two 3x3 convolutional layers with 8 filters each and strides 2 and 1 respectively; each followed by a ReLU nonlinearity except the last one. The outputs are flattened and fed to a 3-layer feed-forward network with 64 hidden units and ReLU nonlinearities, and a final layer with a single neuron. We use dropout in the feed-forward network, with a probability of 0.1. Precedence probabilities are obtained by applying the sigmoid function to the outputs of the last feed-forward layer.
The precedence estimator is trained offline on 100k trajectories collected from a random agent. It is trained on a total of 20M pairs of observations sampled with a window of size 15, although we observed identical performance with larger sizes (up to 120, which is the maximal window size). We use the Adam optimizer with a learning rate of 0.0005, a momentum decay of 0.9, a variance decay of 0.999. We also use weight decay, with a coefficient of 0.0001. We use a threshold $\beta$ of 0.9. We selected hyperparameters based on performance on validation data.

\subsection{Reversibility-Aware Control in Cartpole+}
\label{appendix:cartpole_safe}

\paragraph{Learning $\psi$.}\looseness=-1
The model architecture is the same as described in Appendix~\ref{appendix:reward_free_rl}. The training is done offline using a buffer of 100k trajectories collected using a random policy. State pairs are fed to the classifier in batches of size 128, for a total of 3M pairs. We use the Adam optimizer with a learning rate of 0.01. We use a window $w$ equal to 200, which is the maximum number of timesteps in our environment.

\paragraph{Learning $\phi$.}\looseness=-1
We use a shallow feed-forward network with a single hidden layer of size 64 followed by a ReLU activation. From the same buffer of trajectories used for $\psi$, we sample 100k transitions and feed them to $\phi$ in batches of size 128. As before, training is done using Adam and a learning rate of 0.01.

\subsection{DQN and M-DQN in Cartpole+}
\label{appendix:cartpoleDQN}


We use the same architecture for DQN and M-DQN. The network is a feed-forward network composed of two hidden layers of size 512 followed by ReLU activation. In both cases, we update the online network every 4 timesteps, and the target network every 400 timesteps. We use a replay buffer of size 50k, and sample batches of size 128. We use the Adam optimizer with a learning rate of 0.001.

We train both algorithms for 2M timesteps. We run an evaluation episode every 1000 timesteps, and report the maximum performance encountered during the training process. We perform a grid search for the discount factor $\gamma \in [0.99, 0.999, 0.9997]$, and for M-DQN parameters $\alpha \in [0.7, 0.9, 0.99]$ and $\tau \in [0.008, 0.03, 0.1]$. The best performances were obtained for $\alpha=0.9, \tau = 0.03$, and $\gamma=0.99$.

\subsection{Reversibility-Aware Control in Turf}
\label{appendix:grassland_safe}

\paragraph{Learning $\psi$.}
We use the same model architecture as in RAE (Appendix~\ref{appendix:learning_reversible_policies}), and the same offline training procedure that was used for Cartpole+ (Appendix~\ref{appendix:cartpole_safe}). The window $w$ was set to 120, which is the maximum number of steps in Turf.

\paragraph{Learning $\phi$.}
The architecture is similar to $\psi$, except for the last linear layers: the output of the convolutional layers is flattened and fed to a feed-forward network with one hidden layer of size 64 followed by a ReLU. Again, we used the exact same training procedure as in the case of Cartpole+ (Appendix~\ref{appendix:cartpole_safe}).

\subsection{Safety and Performance Trade-off in Turf}
\label{appendix:grassland_tradeoff}

\begin{figure}
\centering
\begin{subfigure}{.5\textwidth}
  \centering
  \includegraphics[width=1\linewidth]{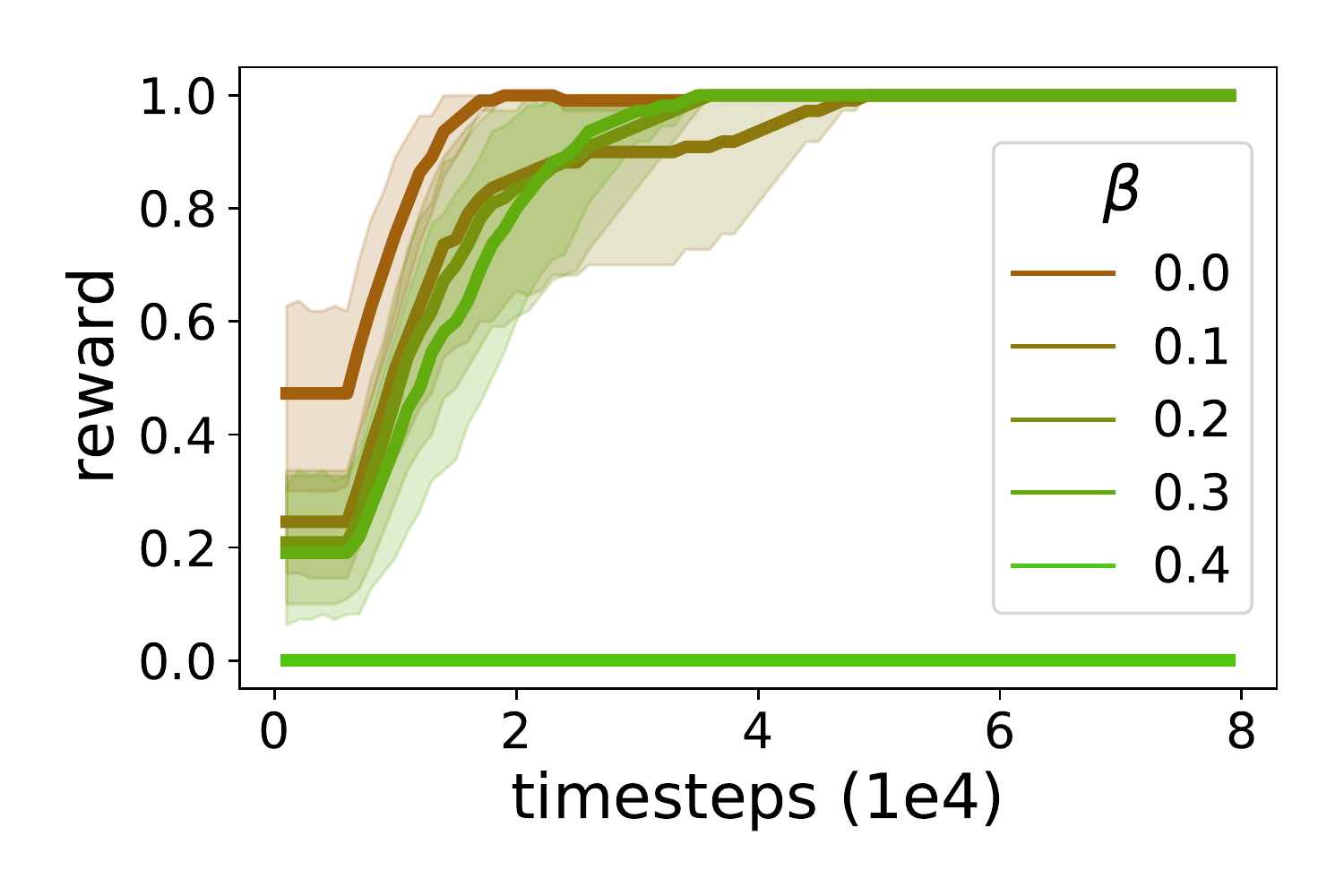}
  \caption{}
  \label{fig:grassland_beta_rewards}
\end{subfigure}%
\begin{subfigure}{.5\textwidth}
  \centering
  \includegraphics[width=1\linewidth]{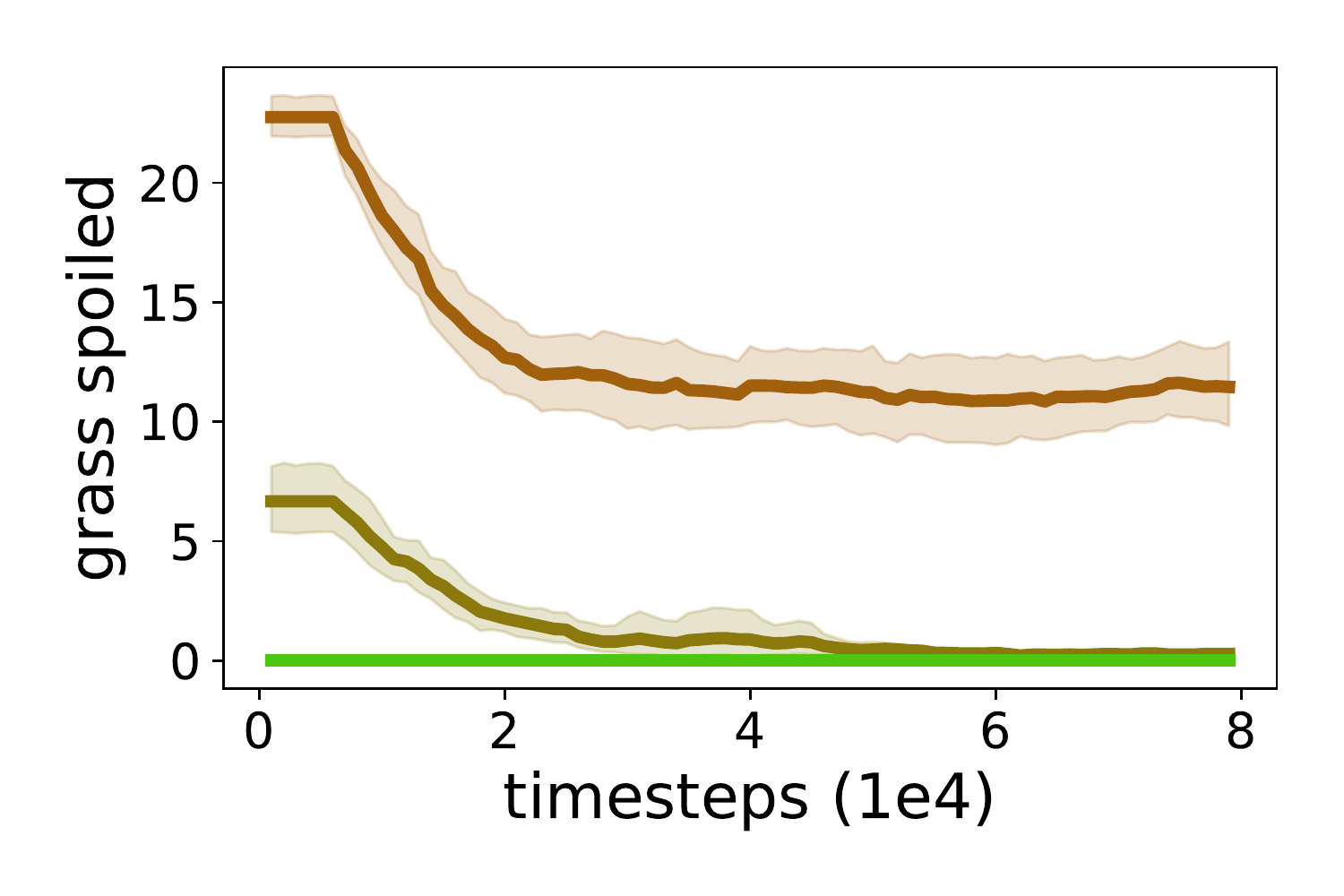}
  \caption{}
  \label{fig:grassland_beta_ruined}
\end{subfigure}
\caption{\textbf{(a):} Reward learning curve for PPO+RAC and several thresholds $\beta$ (average over 10 random seeds). A threshold of 0 means actions are never rejected, and corresponds to the standard PPO. \textbf{(b):}  Number of irreversible side-effects (grass pixels stepped on). For $\beta$ between 0.2 and 0.4, 0 side-effects are induced during the whole learning.}
\label{fig:grassland_beta}
\end{figure}

We investigate the performance-to-safety trade-off induced by reversibility-awareness in Turf. In Fig.~\ref{fig:grassland_beta_rewards}, we see that the agent is not able to reach the goal when the threshold is greater than 0.4: with too high a threshold, every action leading to the goal could be rejected. We also see that it solves the task under lower threshold values, and that lowering $\beta$ results in faster learning. On the other hand, Fig.~\ref{fig:grassland_beta_ruined} shows that achieving zero irreversible side-effects during the learning is only possible when $\beta$ is greater than 0.2. In this setting, the optimal thresholds are thus between 0.2 and 0.3, allowing the agent to learn the new task while eradicating every side-effect.

This experiment gives some insights on how to tune $\beta$ in new environments. It should be initialized at 0.5 and decreased progressively, until the desired agent behaviour is reached. This would ensure that the chosen threshold is the maximal threshold such that the environment can be solved, while having the greatest safety guarantees.

\section{Performance in Stochastic MDPs}

To study how reversibility-awareness helps in stochastic MDPs, we use a 2D cliff-walking gridworld where stochasticity comes from the wind: additionally to its move, the agent is pushed towards the cliff with a fixed probability. The agent gets a +1 reward for each timestep it stays alive, with a maximum of 250 timesteps. A reversibility-aware agent with a well calibrated threshold should avoid most moves that push it towards the cliff.
We use a 6x8 grid, with a maximum of 250 timesteps per episode, and report results averaged over 5000 runs.
We provide two tables: Table~\ref{table:random} with the average scores of a random policy and Table~\ref{table:random-rac} with the average scores of a random policy equipped with RAC. Rows correspond to varying stochasticity and columns to varying threshold values.
\begin{table}[hbt]
\centering
\caption{Scores for a random policy in the 2D cliff-walking gridworld, where $p$ is the probability of being pushed by the wind. Higher is better.}
\begin{tabular}{llllll}
\toprule
$p$ \textbackslash \, Threshold & 0.   & 0.1  & 0.2  & 0.3  & 0.4  \\
\midrule
0.                         & 57.5 & 57.7 & 61.2 & 58.2 & 57.7 \\
0.1                        & 29.8 & 28.8 & 29.5 & 30.2 & 29.6 \\
0.2                        & 18.6 & 18.5 & 19.3 & 18.9 & 18.8 \\
0.3                        & 13.4 & 13.3 & 13.9 & 13.6 & 13.4 \\
0.4                        & 10.5 & 10.7 & 10.4 & 10.2 & 10.2 \\
\end{tabular}
\label{table:random}
\end{table}

\begin{table}[hbt]
\centering
\caption{Scores for a random policy with RAC in the 2D cliff-walking gridworld, where $p$ is the probability of being pushed by the wind. Higher is better.}
\begin{tabular}{llllll}
\toprule
$p$ \textbackslash \, Threshold & 0.   & 0.1   & 0.2   & 0.3   & 0.4   \\
\midrule
0.                         & 59.1 & 250.0 & 250.0 & 250.0 & 250.0 \\
0.1                        & 29.2 & 56.0  & 56.3  & 80.2  & 248.5 \\
0.2                        & 18.7 & 26.7  & 29.2  & 85.8  & 238.6 \\
0.3                        & 13.2 & 16.8  & 19.6  & 77.6  & 250.0 \\
0.4                        & 10.4 & 12.5  & 24.9  & 152.2 & 250.0 \\
\end{tabular}
\label{table:random-rac}
\end{table}

We can notice that:
\begin{itemize}
    \item RAC significantly improves performance (reaching maximum or near-maximum score),
    \item a well-tuned threshold value is crucial to get decent performance with RAC,
    \item the optimal threshold increases with the stochasticity of the environment (but seems to quickly converge).
\end{itemize}

\end{document}